\documentclass{article}
\usepackage[preprint]{neurips_2026}
\usepackage[utf8]{inputenc} 
\usepackage[T1]{fontenc}    
\usepackage{hyperref}       
\usepackage{url}            
\usepackage{booktabs}       
\usepackage{multirow}       
\usepackage{array}          
\usepackage{makecell}       
\usepackage{float}          
\usepackage{graphicx}       
\usepackage{pifont}         
\usepackage{amsfonts}       
\usepackage{float}
\usepackage{nicefrac}       
\usepackage{microtype}      
\usepackage{amsmath}
\usepackage{xcolor}         
\usepackage{listings}
\usepackage{tikz}
\usepackage[table]{xcolor}
\definecolor{oursbg}{RGB}{235,235,255}
\usepackage{pgfplots}
\pgfplotsset{compat=1.18}
\usepackage{xcolor}
\newcolumntype{C}[1]{>{\centering\arraybackslash}m{#1}}
\newcommand{\cmark}{\ding{51}}
\newcommand{\xmark}{\ding{55}}
\title{MMVIAD: Multi-view Multi-task Video Understanding for Industrial Anomaly Detection}

\author{%
  \textbf{Xiran Zhao}$^{1}$\thanks{Equal contribution.} \quad
  \textbf{Jing Jin}$^{2,3}$\footnotemark[1] \quad
  \textbf{Yan Bai}$^{3}$\footnotemark[1] \quad
  \textbf{Zhongan Wang}$^{1}$ \quad
  \textbf{Yifeng Sun}$^{1}$ \\[0.ex] 
  \bfseries Yihang Lou$^{4}$ \quad
  \bfseries Xuanyu Zhu$^{4}$ \quad
  \bfseries Tao Feng$^{2}$ \quad
  \bfseries Yingna Wu$^{1,\dagger}$\thanks{Corresponding author.} \\
  \vspace{0.2em} \\
  \normalsize $^{1}$ShanghaiTech University \quad $^{2}$Tsinghua University \quad $^{3}$Meituan Inc. \quad $^{4}$Peking University \\
  \texttt{\{zhaoxr2025, sunyf12025, wuyn\}@shanghaitech.edu.cn} \\
  \texttt{jingjin0007@gmail.com} \quad \texttt{yanbai02@meituan.com}
}
\begin{document}

\maketitle

\begin{abstract}
Industrial anomaly detection is critical for manufacturing quality control, yet existing datasets mainly focus on static images or sparse views, which do not fully reflect continuous inspection processes in real industrial scenarios. We introduce \textbf{MMVIAD} (Multi-view Multi-task Video Industrial Anomaly Detection), to the best of our knowledge \textbf{the first continuous multi-view video dataset} for industrial anomaly detection and understanding, together with a benchmark for multi-task evaluation. MMVIAD contains object-centric 2-second inspection clips with approximately 120 degrees of camera motion, covering 48 object categories, 14 environments, and 6 structural anomaly types. It supports anomaly detection, defect classification, object classification, and anomaly visible-time localization. Systematic evaluations on MMVIAD show that current commercial and open-source video MLLMs remain far below human performance, especially for fine-grained defect recognition and temporal grounding. To improve transferable anomaly understanding, we further develop a two-stage post-training pipeline where \textbf{PS-SFT} (Perception-Structured Supervised Fine-Tuning) initializes perception-structured reasoning and \textbf{VISTA-GRPO} (Visibility-grounded Industrial Structured Temporal Anomaly Group Relative Policy Optimization) refines the model with semantic-gated defect reward and visibility-aware temporal reward, producing the final model \textbf{VISTA}. On MMVIAD-Unseen, VISTA improves the base model's average score across the four tasks from 45.0 to 57.5, surpassing GPT-5.4.
Source code is available at \url{https://github.com/Georgekeepmoving/MMVIAD}.
\end{abstract}

\section{Introduction}
\begin{figure}[H]
    \centering
    \includegraphics[width=\textwidth]{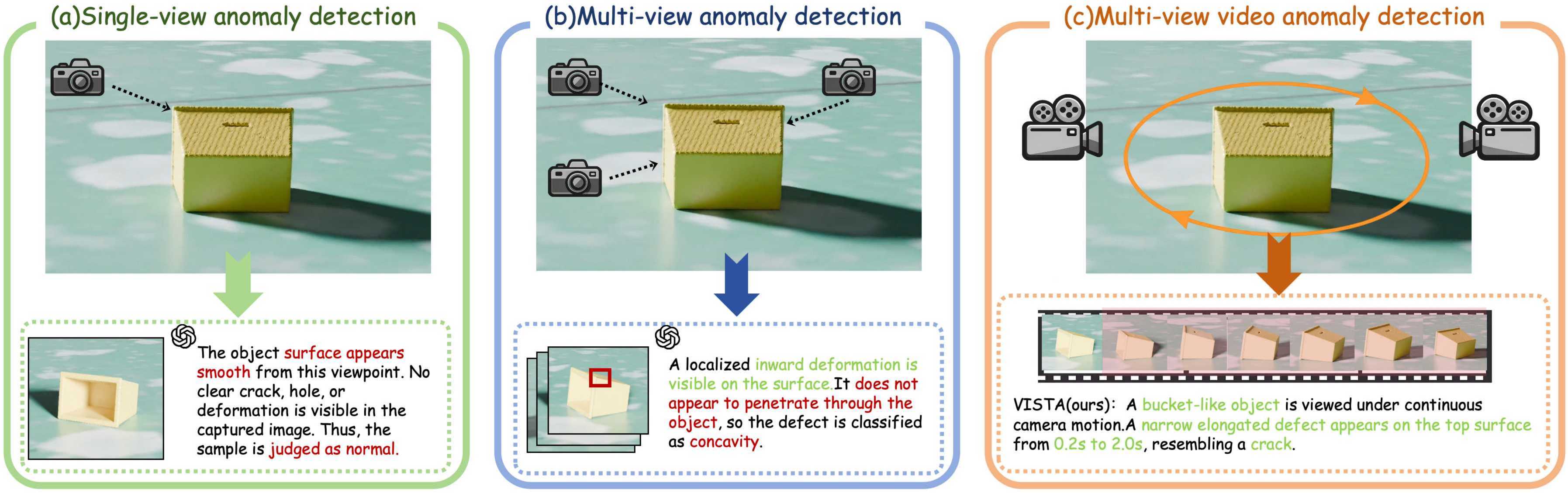}
    \caption{\textbf{Comparison of anomaly detection paradigms.} From single-view and sparse multi-view detection to continuous multi-view video anomaly detection.}
    \label{fig:teaser}
\end{figure}

Industrial visual inspection is rarely limited to a single static view~\cite{mvtecad, realiad, rad}. In practical inspection, a camera may move around a part, or the part may rotate on a fixture, so that different surfaces are observed over a short video. For structural defects such as cracks, holes, bulges, broken regions, scratches, and concavities, the diagnostic evidence is often viewpoint-dependent: a defect can be invisible from one view, weakly visible from another, and clearly diagnosable only during a short interval of the inspection trajectory~\cite{mvtecad, realiad, rad}. Figure~\ref{fig:teaser} illustrates this shift from single-view and sparse multi-view anomaly detection to continuous multi-view video anomaly detection, where models must not only recognize anomaly existence, defect type, and object category, but also identify when the video provides visual evidence for the defect~\cite{timechat, moment_detr, time_r1}.

Existing industrial anomaly benchmarks cover important parts of this problem but do not directly evaluate this continuous multi-view visibility setting, as contrasted in Figure~\ref{fig:teaser}. MVTec AD and VisA established static image-level and pixel-level evaluation~\cite{mvtecad, visa}. MVTec 3D-AD, PAD, Real-IAD, and RAD extend inspection to 3D sensing, pose variation, or sparse multi-view observations~\cite{mvtec3dad, pad, realiad, rad}. Recent video datasets such as Phys-AD further introduce temporal industrial anomaly scenarios~\cite{physad}. However, these datasets generally do not provide continuous multi-view inspection videos with annotations of when viewpoint-dependent structural evidence becomes visible. As a result, a model may predict the correct label while remaining ungrounded in the diagnostic video interval.

To address this gap, we introduce \textbf{MMVIAD} (Multi-view Multi-task Video Industrial Anomaly Detection), a visibility-grounded video dataset and benchmark for continuous multi-view, multi-task industrial anomaly understanding. To the best of our knowledge, MMVIAD is the first continuous multi-view video dataset for industrial anomaly detection. MMVIAD is built from controllable object-centric rendering trajectories, where each 2-second clip covers approximately 120 degrees of viewpoint change~\cite{anomaly_shapenet, pad}. We use controllable rendering because accurate visible-time annotation requires aligned viewpoints and verifiable defect visibility, which are difficult to obtain reliably from real inspection videos. The dataset contains over four thousand inspection clips across 48 object categories, 14 environments, and 6 structural anomaly types. For each anomalous object, we generate two aligned videos under identical rendering conditions: an anomaly-unmarked video that renders the anomalous object without additional visual cues, and an anomaly-marked video where the defective region is highlighted in red for visibility annotation. By comparing the aligned videos and manually verifying the candidate intervals, MMVIAD obtains visible-time annotations for four coupled QA-style tasks: anomaly detection, defect classification, object classification, and anomaly visible-time localization.

These four tasks are deliberately coupled. Defect classification is meaningful only after anomaly detection is correct; object identity provides geometric and semantic context for interpreting the defect; visible-time localization asks the model to bind the predicted defect to the temporal interval where its evidence appears. MMVIAD therefore evaluates whether a video MLLM can connect object-level perception, defect semantics, and temporal grounding, rather than solve four independent classification problems. To provide a reference training baseline for this setting, we further study a two-stage structured post-training pipeline. The first stage, \textbf{PS-SFT} (Perception-Structured Supervised Fine-Tuning), follows multimodal instruction tuning to teach the model to separate whole-clip perception from localized temporal evidence~\cite{llava, instructblip}. The second stage, \textbf{VISTA-GRPO} (Visibility-grounded Industrial Structured Temporal Anomaly Group Relative Policy Optimization), refines the model with task-aware rewards inspired by recent rule-based reinforcement learning, including a semantic gate for defect classification and a visibility-aware reward for temporal localization~\cite{deepseek_r1, video_r1, time_r1}. The resulting model \textbf{VISTA} serves as our reference post-training baseline.

We evaluate humans, commercial video MLLMs, and open-source video MLLMs under two MMVIAD protocols~\cite{qwen3_vl, video_r1, time_r1, videochat_r1, videochat_r15}. MMVIAD-Standard measures overall performance when train and test clips cover the same object categories, while MMVIAD-Unseen tests whether training transfers to unseen object categories. The results show a persistent gap between current models and human annotators, especially on fine-grained defect classification and anomaly visible-time localization. On MMVIAD-Unseen, VISTA improves the Qwen3-VL-8B base model from 45.0 to 57.5 average score, with the largest gains in anomaly detection, temporal localization, and object classification. These results suggest that MMVIAD exposes a failure mode of current video MLLMs: they can often recognize coarse object or anomaly cues, but still struggle to bind defect semantics to the viewpoint interval where the evidence is available.We summarize our contributions as follows:
\begin{itemize}
    \item \textbf{Dataset and benchmark.} We introduce MMVIAD, to the best of our knowledge the first continuous multi-view video dataset for industrial anomaly detection, with QA-based tasks and visible-time annotations for viewpoint-dependent structural defects.
    \item \textbf{Diagnostic evaluation.} We evaluate human annotators, commercial models, and open-source video MLLMs on MMVIAD-Standard and MMVIAD-Unseen, showing that defect semantics and visible-time grounding remain key bottlenecks.
    \item \textbf{Reference post-training.} We propose VISTA, a two-stage post-training baseline that combines PS-SFT initialization with VISTA-GRPO reward-based refinement to improve generalization on MMVIAD-Unseen.
\end{itemize}
\section{Related Work}
\textbf{Industrial Anomaly Detection.}
Industrial anomaly detection has been shaped by static-image and sparse-view benchmarks. MVTec AD, MPDD, and VisA established image-level and pixel-level evaluation~\cite{mvtecad, mpdd, visa}, while MVTec 3D-AD, PAD, Real-IAD, and RAD extended the setting toward 3D sensing, pose variation, and multi-view observation~\cite{mvtec3dad, pad, realiad, rad}. Recent video datasets such as Phys-AD further extend industrial anomaly detection to video scenarios~\cite{physad}. In parallel, methods such as PaDiM, CutPaste, DRAEM, Reverse Distillation, PatchCore, and EfficientAD advanced distribution modeling, self-supervised reconstruction, memory-bank, and distillation-based anomaly detection~\cite{padim, cutpaste, draem, reverse_distillation, patchcore, efficientad}. Recent work further connects industrial inspection with multimodal large models and reasoning-oriented post-training~\cite{anomalygpt,mmad,anomalyr1}. Despite this progress, existing benchmarks still mainly evaluate static images, pixel maps, sparsely sampled views, or single-view videos, rather than continuous multi-view videos where anomaly evidence may be temporally visible only under certain viewpoints. MMVIAD targets this missing setting.

\textbf{Video Understanding and Anomaly Visible-time Localization.}
Temporal video grounding localizes moments from language queries, from early methods such as DiDeMo/MCN and TALL to temporal-map and transformer-based models such as 2D-TAN and Moment-DETR~\cite{localizing_moments, tall, two_d_tan, moment_detr}. Video LMMs further expanded instruction-following video understanding~\cite{videollama, video_chatgpt, llama_vid, videollama2, longva}, while timestamp-aware models and reinforcement-style video reasoning methods improve temporal grounding with explicit time or verifiable objectives~\cite{timechat, vtg_llm, trace, chatvtg, video_r1, time_r1, videochat_r1, videochat_r15}. However, these works mainly target natural videos, human actions, or language-described events. They do not address industrial structural defects whose evidence is subtle, viewpoint-dependent, and tied to objectively visible intervals under controlled inspection. Our work adapts anomaly visible-time localization to anomaly visibility grounding in industrial videos.

\textbf{Multimodal Instruction Tuning and Reinforcement Learning.}
Multimodal instruction tuning aligns visual encoders with large language models for instruction-following visual reasoning~\cite{flamingo, blip2, llava, instructblip, minigpt4, qwen_vl}. Recent post-training methods, including chain-of-thought, preference optimization, and policy-gradient reinforcement learning, further improve structured reasoning through imitation or reward-based objectives~\cite{cot, instructgpt, dpo, deepseekmath, deepseek_r1}. However, these techniques have not been systematically studied for continuous multi-view industrial anomaly understanding with coupled defect recognition and anomaly visible-time localization.

\section{Method}
\subsection{MMVIAD Dataset Construction}
\label{sec:dataset_construction}
\begin{figure}[H]
    \centering
    \includegraphics[width=\textwidth]{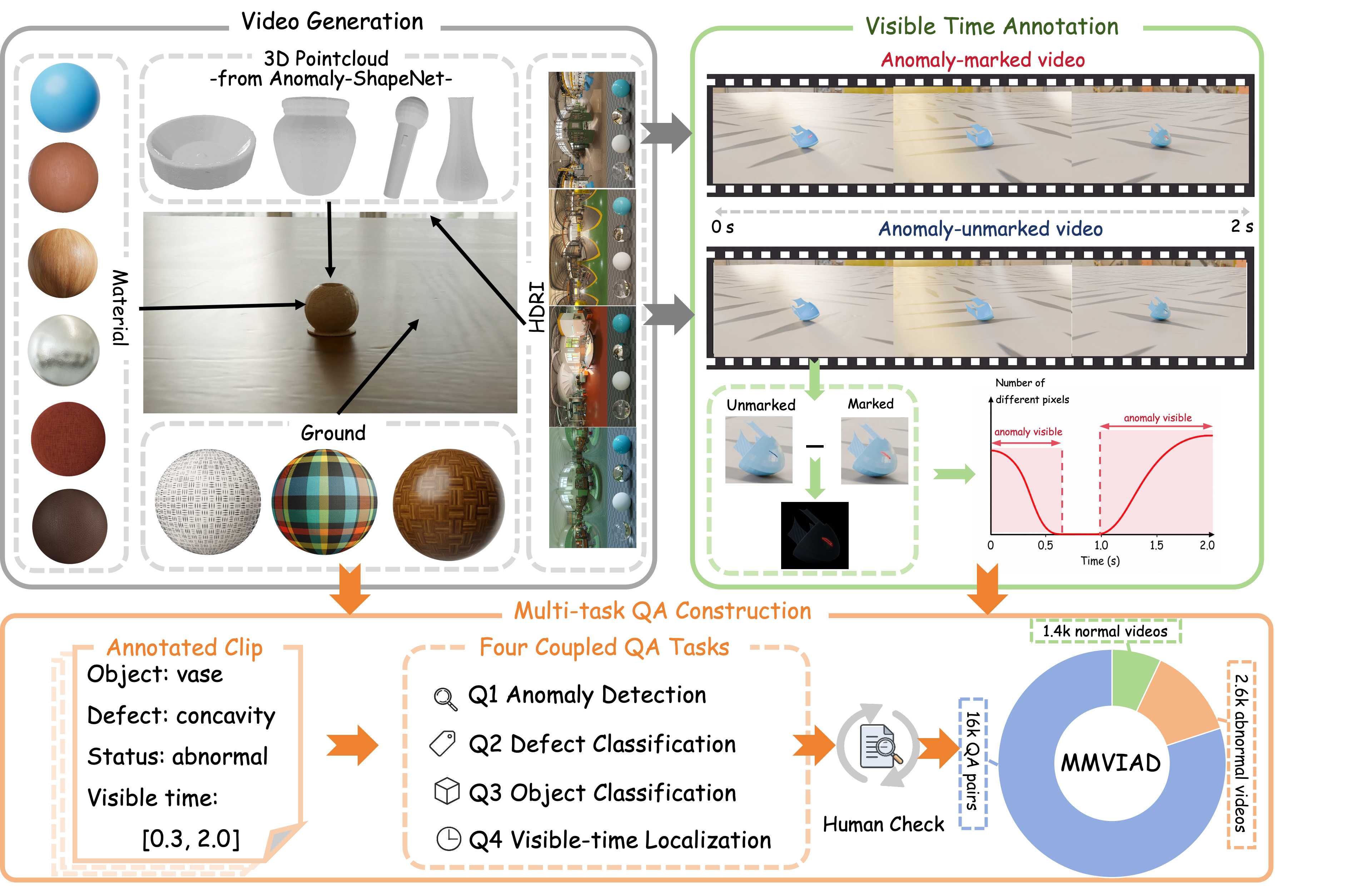}
    \caption{\textbf{MMVIAD data construction.}
We render continuous multi-view videos from Anomaly-ShapeNet point clouds, derive visible-time labels by comparing aligned marked and unmarked videos, and convert each annotated clip into four QA tasks for anomaly detection, defect classification, object classification, and visible-time localization.}
\label{fig:data_part}
\end{figure}

We construct MMVIAD for multi-task, multi-view industrial anomaly understanding in continuous
inspection videos. A key challenge is obtaining precise and verifiable labels for when structural
anomaly evidence is visible. As illustrated in Figure~\ref{fig:data_part}, MMVIAD is constructed through three stages: controllable video generation, visible-time annotation, and multi-task QA construction. Their frame-wise comparison provides candidate visible intervals,
which are then manually verified to ensure that the final labels reflect the visibility of structural
anomalies rather than illumination, material, or background changes. This design reduces the ambiguity
and subjectivity of manually judging defect visibility in continuous videos. Table~\ref{tab:dataset_comparison}
shows that prior industrial anomaly benchmarks are largely image-based and do not jointly support
continuous viewpoint variation and QA-based anomaly understanding.

\begin{table}[H]
\centering
\caption{\textbf{Comparison of industrial anomaly detection datasets.}
Repr. denotes representation, Type denotes data type, Class denotes the number of object categories,
Cont. denotes continuous viewpoint variation, and QA denotes question-answering-based anomaly understanding.}
\label{tab:dataset_comparison}
{\fontsize{7.3}{9.0}\selectfont
\setlength{\tabcolsep}{1.5pt}
\renewcommand{\arraystretch}{0.85}
\begin{tabular*}{\textwidth}{@{\extracolsep{\fill}}
C{1.55cm} C{1.05cm} C{0.75cm} C{0.65cm} C{0.55cm}
C{0.85cm} C{0.90cm} C{0.85cm} C{0.55cm} C{0.50cm}}
\toprule
Dataset & Venue & Repr. & Type & Class & Normal & Abnormal & View & Cont. & QA \\
\midrule
MVTec AD \cite{mvtecad} & CVPR'19 & image & real  & 15 & 4,096  & 1,258  & single & \xmark & \xmark \\
VisA \cite{visa}        & ECCV'22 & image & real  & 12 & 9,621  & 1,200  & single & \xmark & \xmark \\
PAD \cite{pad}          & NeurIPS'23 & image & synth & 20 & 4,960  & 4,412  & $\sim$20 & \xmark & \xmark \\
Real-IAD \cite{realiad} & CVPR'24 & image & real  & 30 & 99,721 & 51,329 & 5      & \xmark & \xmark \\
RAD \cite{rad}          & arXiv'24 & image & real  & 13 & 1,224  & 3,063  & 68     & \xmark & \xmark \\
Phys-AD \cite{physad}   & CVPR'25 & video & real  & 49 & 3,598  & 2,836  & single & \xmark & \xmark \\
\midrule
\textbf{MMVIAD} & - & \textbf{video} & synth & 48 &
1,410 & 2,613 & \textbf{120$^\circ$} & \cmark & \cmark \\
\bottomrule
\end{tabular*}
}
\end{table}

\textbf{Video generation.}
MMVIAD is built from object-centric 360$^\circ$ rendering trajectories. We use controllable rendering because anomaly visible-time localization requires frame-level visibility labels that are difficult to obtain reliably from real captured videos. The objects used in our videos are sourced from Anomaly-ShapeNet~\cite{anomaly_shapenet}, a ShapeNet-based synthetic 3D anomaly dataset constructed through mesh subdivision and defect carving. We render each object in Blender at 1920$\times$1080 and 30 fps, while randomizing floor appearance, HDRI illumination, material properties, camera viewpoints, and object poses to improve visual diversity. Although rendered videos do not fully replace real industrial acquisition, they allow precise control of viewpoint trajectories, object pose, illumination, and defect geometry, making it possible to generate repeatable inspection clips and verifiable visible-time annotations. For each anomalous object, we generate two aligned videos under identical rendering conditions: one anomaly-unmarked video and one anomaly-marked video where the defective region is highlighted in red. Each complete 360$^\circ$ trajectory is then split into 2-second clips, and each clip is treated as an independent sample.

\paragraph{Visible-time annotation.}
As shown in Figure~\ref{fig:data_part}, we annotate anomaly visibility by comparing two aligned videos for each anomalous sample. The first is an anomaly-unmarked video, where the anomalous object is rendered normally without visual highlighting. The second is an anomaly-marked video, where the defective region is highlighted in red. Both videos are rendered under identical camera motion, illumination, material, floor, and object-pose conditions. We compare the two videos frame by frame to obtain candidate intervals in which the marked defective region is visible. These candidate intervals are then manually checked and refined to ensure that the final label corresponds to the time span where the structural anomaly itself is visible, rather than visual differences caused by lighting, material reflectance, pose variation, or background interference. The resulting annotation is a visible-time interval $[t_{\mathrm{start}}, t_{\mathrm{end}}]$, which serves as supervision for anomaly visible-time localization.

\textbf{QA-pair generation.}
Industrial video anomaly understanding requires solving four coupled sub-tasks: binary anomaly detection (Q1), conditionally dependent defect categorization (Q2), object classification (Q3), and anomaly visible-time localization (Q4). MMVIAD covers six structural anomaly types, including crack, scratch, concavity, bulge, broken, and hole. For Q1--Q3, we construct standardized question-answer pairs using the normal/abnormal labels, defect-type labels, and object-category labels from Anomaly-ShapeNet. For Q4, we use the visible-time annotations described above as supervision and ask models to output the time interval in which diagnostic anomaly evidence is visible, in the format $[\texttt{start\_sec}, \texttt{end\_sec}]$. With unified question templates and answer formats, MMVIAD places anomaly existence, defect semantics, object semantics, and anomaly visible-time localization under a single evaluation interface. Finally, each video clip is converted into four aligned QA instances, yielding 16,092 QA pairs in total.

\subsection{PS-SFT Initialization and VISTA-GRPO Refinement}

The four tasks in MMVIAD are intrinsically coupled. Defect classification is meaningful only when an anomaly is correctly detected, object classification provides semantic context for anomaly understanding, and anomaly visible-time localization requires grounding defect evidence in temporal intervals. These dependencies make naive multi-task optimization susceptible to cross-task interference, reward sparsity, and unstable structured outputs. We therefore adopt a two-stage training pipeline. PS-SFT first initializes the model with perception-structured reasoning traces, and VISTA-GRPO then refines the policy with a semantically gated multi-task reward and group-relative policy optimization.

\begin{figure}[H]
    \centering
    \includegraphics[width=\textwidth]{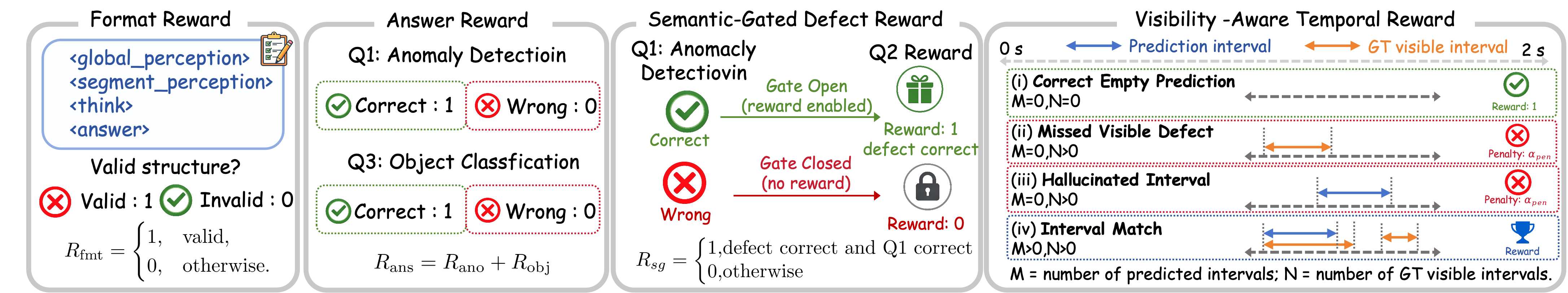}
    \caption{\textbf{VISTA-GRPO reward modeling.} The reward design combines format checking, answer correctness, semantic-gated defect reward, and visibility-aware temporal reward.}
    \label{fig:reward_modeling}
\end{figure}
\subsubsection{PS-SFT: Perception-Structured Supervised Fine-Tuning}

Directly applying reinforcement learning to the base model is unstable, as unreliable structured outputs lead to sparse rewards and hinder early exploration. To mitigate this cold-start problem, we first construct \textbf{PS-SFT} to initialize the model with perception-structured reasoning behavior before reinforcement learning.

PS-SFT is a teacher-synthesized SFT dataset derived from the MMVIAD training split. For each training clip, we provide Gemini~3.1~Pro with ground-truth labels for all four tasks and instruct it to generate a reasoning trace in the target format: \texttt{<global\_perception>}, \texttt{<segment\_perception>}, \texttt{<think>}, and \texttt{<answer>}. This format decomposes each response into whole-clip perception, localized defect evidence with temporal visibility cues, brief reasoning, and final structured answers. Since generation is conditioned on ground-truth labels, the teacher mainly converts existing annotations into perception-decomposed reasoning traces rather than re-annotating the data.

We discard outputs that miss required sections, violate the section order, or contain unparsable final answers. The remaining traces form $\mathcal{D}_{\mathrm{PS\text{-}SFT}}$, the perception-structured SFT dataset used for supervised initialization. We then fine-tune the base model on $\mathcal{D}_{\mathrm{PS\text{-}SFT}}$ with teacher forcing by minimizing the standard causal language modeling loss. Given a video $v$, task prompt $\mathcal{P}$, and target output sequence $o=\{o_t\}_{t=1}^{T}$, the PS-SFT objective is:
\begin{equation}
    \mathcal{L}_{\mathrm{PS\text{-}SFT}}
    = -\sum_{t=1}^{T}
      \log p_{\theta}\!\left(o_t \mid v,\,\mathcal{P},\,o_{<t}\right).
\end{equation}
PS-SFT is not intended to maximize final task performance by itself; instead, it provides a stable initialization for subsequent VISTA-GRPO refinement. It aligns the output format, encourages perception decomposition, and primes the model with initial anomaly awareness and temporal grounding before reward-driven optimization.

As illustrated in Figure~\ref{fig:reward_modeling}, VISTA-GRPO combines four reward components: format checking, answer correctness, semantic-gated defect reward, and visibility-aware temporal reward.
\paragraph{Format and Answer Rewards.}
The format reward preserves the perception-structured output format learned during PS-SFT, while the answer reward evaluates Q1 anomaly detection and Q3 object classification:
\begin{align}
R_{\mathrm{fmt}}(o)
&= \mathbf{1}\!\left[o \text{ follows the required structure}\right], \\
R_{\mathrm{ans}}(o)
&= \mathbf{1}\!\left[\hat{y}_{\mathrm{ano}}=y_{\mathrm{ano}}\right]
 + \mathbf{1}\!\left[\hat{y}_{\mathrm{obj}}=y_{\mathrm{obj}}\right].
\end{align}
The two terms in $R_{\mathrm{ans}}$ correspond to Q1 and Q3, and the object reward is computed unconditionally because object identity is independent of defect status.

\paragraph{Semantic-Gated Defect Reward.}
Defect classification corresponds to Q2, but it is not an independent classification task. In industrial inspection, predicting a defect type is meaningful only when the model first correctly detects that an anomaly exists. A plain defect classification reward may encourage the policy to exploit defect-class priors even when Q1 is incorrect. We therefore introduce a semantic-gated defect reward:
\begin{equation}
R_{\mathrm{sg}}(o)=
\begin{cases}
1, & \text{if } \hat{y}_{\mathrm{def}} = y_{\mathrm{def}}
     \land \hat{y}_{\mathrm{ano}} = y_{\mathrm{ano}},\\
0, & \text{otherwise}.
\end{cases}
\label{eq:semantic_gate}
\end{equation}
This reward explicitly enforces the Q1$\rightarrow$Q2 dependency, making defect classification contingent on valid anomaly awareness and reducing reward hacking from defect-class priors.

\paragraph{Visibility-Aware Temporal Reward.}
Anomaly visible-time localization corresponds to Q4. Unlike ordinary temporal localization, this task requires the model to determine both whether visible anomaly evidence exists and when it appears. A flat IoU reward conflates different cases, such as correctly predicting an empty interval for a normal sample, missing a visible defect, and hallucinating an interval on a defect-free sample. We therefore design a visibility-aware temporal reward that decouples visibility presence from boundary precision.

Let $\hat{\mathcal{I}}=\{[\hat{t}_{s}^{(k)},\hat{t}_{e}^{(k)}]\}_{k=1}^{M}$ denote the set of predicted visible-time intervals, where $M$ is the number of predicted intervals. Let $\mathcal{I}^{*}=\{[t_{s}^{(k)},t_{e}^{(k)}]\}_{k=1}^{N}$ denote the set of ground-truth visible-time intervals, where $N$ is the number of annotated intervals. Thus, $M=0$ means that the model predicts no visible anomaly interval, while $N=0$ means that the clip has no annotated visible anomaly interval. The reward is defined as:
\begin{equation}
R_{\mathrm{vis}}(o)=
\begin{cases}
1.0, & M=0 \wedge N=0,\\[3pt]
\alpha_{\mathrm{pen}}, & M=0 \wedge N>0,\\[3pt]
\alpha_{\mathrm{pen}}, & M>0 \wedge N=0,\\[3pt]
\alpha_{\mathrm{bon}}
+\alpha_{\mathrm{iou}}\cdot
\mathrm{IoU}_{\max}(\hat{\mathcal{I}},\mathcal{I}^{*})
\cdot e^{-\lambda |M-N|}, & M>0 \wedge N>0.
\end{cases}
\label{eq:visibility_reward}
\end{equation}
The first case rewards correct empty prediction on normal samples, the second penalizes missed visible defects, and the third penalizes hallucinated intervals. When both prediction and ground truth contain visible intervals, the last case evaluates boundary quality using maximum IoU, a discovery bonus, and an interval-count penalty:
\begin{equation}
\mathrm{IoU}_{\max}(\hat{\mathcal{I}},\mathcal{I}^{*})
=
\max_{\hat{I}\in\hat{\mathcal{I}},\, I^{*}\in\mathcal{I}^{*}}
\frac{|\hat{I} \cap I^{*}|}{|\hat{I} \cup I^{*}|}.
\end{equation}
We set $\alpha_{\mathrm{bon}}=0.3$, $\alpha_{\mathrm{iou}}=0.7$, $\alpha_{\mathrm{pen}}=-0.3$, and $\lambda=0.5$. The discovery bonus provides effective feedback when boundary precision is still poor, while $e^{-\lambda |M-N|}$ discourages over- and under-segmentation.

During training, VISTA-GRPO samples $G$ responses $\{o_i\}_{i=1}^{G}$ for each prompt and computes a rule-based reward $R_{\mathrm{VISTA\text{-}GRPO}}(o_i)$ from the above reward components. The advantage of each response is estimated relative to the reward distribution within the same group:
\begin{equation}
A_i =
\frac{
R_{\mathrm{VISTA\text{-}GRPO}}(o_i)
-
\mathrm{mean}\!\left(\{R_{\mathrm{VISTA\text{-}GRPO}}(o_j)\}_{j=1}^{G}\right)
}{
\mathrm{std}\!\left(\{R_{\mathrm{VISTA\text{-}GRPO}}(o_j)\}_{j=1}^{G}\right)
}.
\label{eq:vista_grpo_advantage}
\end{equation}
The policy is then updated with the standard GRPO objective, using the PS-SFT checkpoint as both the initial policy and the reference model.

\section{Experiments}
\label{sec:experiments}
\subsection{Setup}
\label{sec:exp_setup}

\paragraph{Benchmark protocols.}
We use two MMVIAD protocols. \textbf{MMVIAD-Standard} partitions clips at the sample level over the same 48 fine-grained object classes, with 2,913 training clips and 1,101 test clips, corresponding to 11,652 and 4,404 QA pairs. It is used to compare human annotators, commercial models, and open-source MLLMs. \textbf{MMVIAD-Unseen} trains on 36 object classes and tests on 12 unseen classes, with 2,952 training clips and 1,062 test clips, corresponding to 11,808 and 4,248 QA pairs. It is used to evaluate transferable anomaly understanding with PS-SFT and VISTA-GRPO.
\paragraph{Human evaluation.}
Following MMAD~\cite{mmad}, we sample 621 clips for human evaluation by eight participants: three experts with industrial anomaly detection experience and five ordinary participants. Each participant answers the same four QA tasks with the same visual inputs as machine baselines, and we report group averages.
\paragraph{Training details.}
Unless otherwise specified, we use Qwen3-VL-8B~\cite{qwen3_vl} as the base model. Training uses $1008 \times 560$ frames with 16 uniformly sampled frames per clip, while evaluation uses 32 frames. Training takes about 6 hours on 4 NVIDIA H100 80GB GPUs. All machine baselines share the same task description, decoding setup, and answer-parsing protocol. We report classification accuracy, localization mIoU, and their arithmetic mean as Avg.
\subsection{MMVIAD-Standard Benchmark Results}
\label{sec:main_results}
Since MMVIAD-Standard is designed for zero-shot evaluation, we exclude VISTA from Table~\ref{tab:main_results} because it is trained on MMVIAD and thus not directly comparable to off-the-shelf MLLMs. Table~\ref{tab:main_results} shows that MMVIAD-Standard remains far from saturated, with the strongest machine baseline still more than 20 points behind expert annotators in Avg. Models perform relatively better on coarse anomaly detection and object classification, but struggle with fine-grained defect classification and anomaly visible-time localization, indicating that generic video QA and multimodal reasoning pretraining are insufficient for structural anomaly understanding under continuous viewpoint changes.

\subsection{MMVIAD-Unseen Generalization Results}

Table~\ref{tab:generalization_results} shows that MMVIAD-Unseen remains challenging for all models. VISTA achieves the best overall performance, improving the Qwen3-VL-8B base model from 45.0 to 57.5 Avg., with clear gains in anomaly detection, visible-time localization, and object classification. The smaller improvement in defect classification suggests that fine-grained defect semantics remain difficult under unseen object categories.
\begin{table}[H]
\centering
\caption{\textbf{Results on MMVIAD-Standard.} The four task columns correspond to anomaly detection (Defect Detect.), anomaly classification (Defect Class.), anomaly visible-time localization (Defect Loc.), and object classification (Object Class.). Avg. denotes the arithmetic mean over the four tasks.}
\label{tab:main_results}
{\fontsize{7.3}{9.0}\selectfont
\setlength{\tabcolsep}{1.5pt}
\renewcommand{\arraystretch}{0.83}
\begin{tabular*}{\textwidth}{@{\extracolsep{\fill}} C{1.7cm} C{3.2cm} C{1.1cm} C{0.9cm} ccc c c}
\toprule
\multirow{2}{*}{Type} & \multirow{2}{*}{Model} & \multirow{2}{*}{Venue} & \multirow{2}{*}{Param.}
& \multicolumn{3}{c}{Defect} & \multicolumn{1}{c}{Object} & \multirow{2}{*}{Avg.} \\
\cmidrule(lr){5-7} \cmidrule(lr){8-8}
& & & & Detect. & Class. & Loc. & Class. & \\
\midrule
- & Human (expert)   & -         & -  & 91.2 & 77.8 & 82.3 & 93.7 & 86.3 \\
- & Human (ordinary) & -         & -  & 87.3 & 67.1 & 78.4 & 91.5 & 81.1 \\
\midrule
\multirow{3}{*}{Commercial}
& GPT-5.4  & -  & -  & 64.1 & 37.9 & 50.5 & 73.0 & 56.4 \\
& GPT-5.4-mini  & -  & -  & 68.3  & 32.6 & 46.3 & 64.5 & 52.9  \\
& Gemini 3.1 Pro & -   & -   & 69.3 & 41.6 & 48.5  & 87.8  & 61.8  \\
\midrule
\multirow{6}{*}{Open Source}
& Qwen3-VL \cite{qwen3_vl}                & arXiv'25   & 2B & 33.9 & 12.4 & 20.7 & 51.8 & 29.7 \\
& Qwen3-VL \cite{qwen3_vl}                & arXiv'25   & 4B & 33.9 & 31.0 & 31.9 & 65.8 & 40.7 \\
& Qwen3-VL \cite{qwen3_vl}                & arXiv'25   & 8B & 35.1 & 31.0 & 30.9 & 69.5 & 41.6 \\
& Video-R1 \cite{video_r1}                & NeurIPS'25 & 7B & 41.4 & 30.6 & 31.1 & 60.9 & 41.0 \\
& Time-R1 \cite{time_r1}                  & NeurIPS'25 & 3B & 49.3 & 26.6 & 30.1 & 63.8 & 42.5 \\
& Time-R1 \cite{time_r1}                  & NeurIPS'25 & 7B & 40.8 & 31.3 & 31.1 & 65.0 & 42.1 \\
& VideoChat-R1 \cite{videochat_r1}        & arXiv'25   & 7B & 41.1 & 32.1 & 31.1 & 68.8 & 43.3 \\
& VideoChat-R1.5 \cite{videochat_r15}     & NeurIPS'25 & 7B & 36.7 & 30.9 & 31.1 & 66.2 & 41.2 \\
\bottomrule
\end{tabular*}
}
\end{table}

\begin{table}[H]
\centering
\caption{\textbf{Results on MMVIAD-Unseen.}}
\label{tab:generalization_results}
{\fontsize{7.3}{9.0}\selectfont
\renewcommand{\arraystretch}{0.83}
\resizebox{\textwidth}{!}{%
\begin{tabular}{C{1.7cm} C{3.2cm} C{1.1cm} C{0.9cm} ccc c c}
\toprule
\multirow{2}{*}{Type} & \multirow{2}{*}{Model} & \multirow{2}{*}{Venue} & \multirow{2}{*}{Param.}
& \multicolumn{3}{c}{Defect} & \multicolumn{1}{c}{Object} & \multirow{2}{*}{Avg.} \\
\cmidrule(lr){5-7} \cmidrule(lr){8-8}
& & & & Detect. & Class. & Loc. & Class. & \\
\midrule
- & Human (expert)   & - & - & 88.5 & 78.9 & 80.1 & 96.4 & 86.0 \\
- & Human (ordinary) & - & - & 83.2 & 73.3 & 77.8 & 93.6 & 82.0 \\
\midrule
\multirow{3}{*}{Commercial}
& GPT-5.4        & - & - & 58.2 & 38.0 & 52.0 & 65.7 & 53.5 \\
& GPT-5.4-mini   & - & - & 62.0 & 29.8 & 60.0 & 62.2 & 53.5 \\
& Gemini 3.1 Pro & - & - & 69.1 & 46.9 & 59.3 & 80.3 & 63.9 \\
\midrule
\multirow{7}{*}{Open Source}
& Qwen3-VL \cite{qwen3_vl}               & arXiv'25   & 2B & 32.3 & 26.2 & 27.4 & 54.8 & 35.2 \\
& Qwen3-VL \cite{qwen3_vl}               & arXiv'25   & 4B & 35.2 & 33.5 & 37.8 & 56.9 & 40.9 \\
& Qwen3-VL \cite{qwen3_vl}               & arXiv'25   & 8B & 39.5 & 36.2 & 37.0 & 67.4 & 45.0 \\
& Video-R1 \cite{video_r1}               & NeurIPS'25 & 7B & 49.5 & 31.1 & 48.4 & 57.0 & 46.5 \\
& Time-R1 \cite{time_r1}                 & NeurIPS'25 & 7B & 44.2 & 35.6 & 48.6 & 56.4 & 46.2 \\
& VideoChat-R1 \cite{videochat_r1}       & arXiv'25   & 7B & 41.1 & 34.9 & 48.8 & 56.9 & 45.4 \\
& VideoChat-R1.5 \cite{videochat_r15}    & NeurIPS'25 & 7B & 41.7 & 34.8 & 45.4 & 55.4 & 44.3 \\
\midrule
\rowcolor{oursbg}
\textbf{Ours}
& \textbf{VISTA (Qwen3-VL)} & \textbf{-} & \textbf{8B}
& \textbf{60.7} & \textbf{37.6} & \textbf{49.6} & \textbf{81.9} & \textbf{57.5} \\
\bottomrule
\end{tabular}
}
}
\end{table}

\begin{figure}[H]
    \centering
    \includegraphics[width=1.0\linewidth]{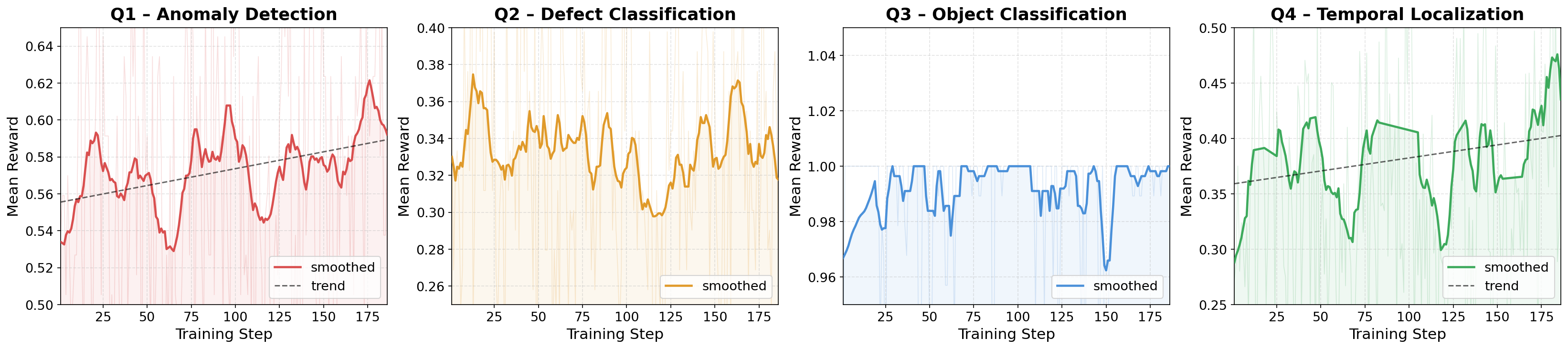}
    \caption{\textbf{Task-wise reward dynamics during VISTA-GRPO training.} Rewards improve most clearly on anomaly detection and anomaly visible-time localization, while defect classification remains sparse and unstable, highlighting the core challenge of fine-grained industrial anomaly understanding.}
    \label{fig:reward_curves}
\end{figure}

Figure~\ref{fig:reward_curves} illustrates the training dynamics. Rewards for anomaly detection and anomaly visible-time localization show clearer upward trends, consistent with Table~\ref{tab:generalization_results}, while defect classification fluctuates more strongly due to sparse seven-way supervision. Object classification remains stable, suggesting that VISTA-GRPO mainly refines anomaly-centric reasoning.

\subsection{Ablation Studies}
\begin{figure}[H]
    \centering
    \includegraphics[width=\textwidth]{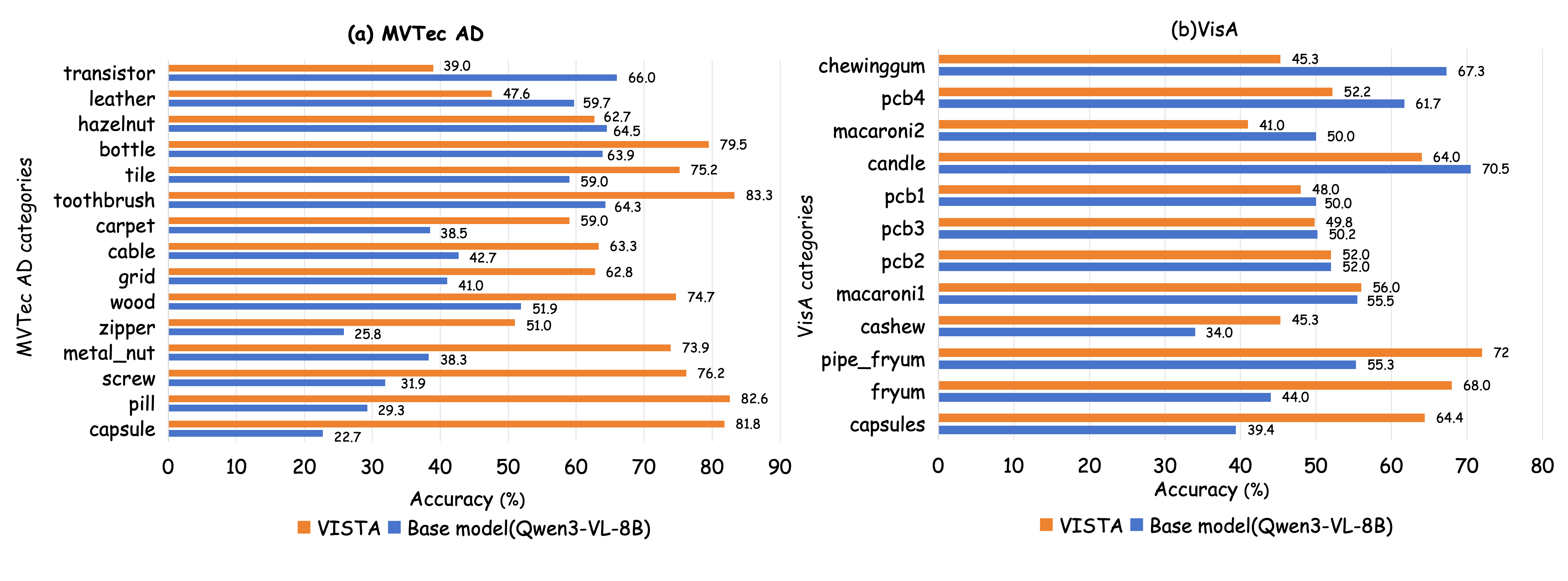}
    \caption{\textbf{Per-category synthetic-to-real generalization.}}
    \label{fig:synthetic_to_real_category}
\end{figure}

Figure~\ref{fig:synthetic_to_real_category} evaluates synthetic-to-real generalization on real industrial image datasets. VISTA trained on synthetic MMVIAD-Unseen improves the Qwen3-VL-8B base model from 43.8 to 67.1 average accuracy on MVTec AD and from 53.0 to 54.4 on VisA. The gain is especially large on MVTec AD, where most categories benefit from the MMVIAD-trained model, suggesting that anomaly understanding learned from synthetic multi-view videos can transfer to real inspection images.

\begin{figure}[H]
    \centering
    \includegraphics[width=\textwidth]{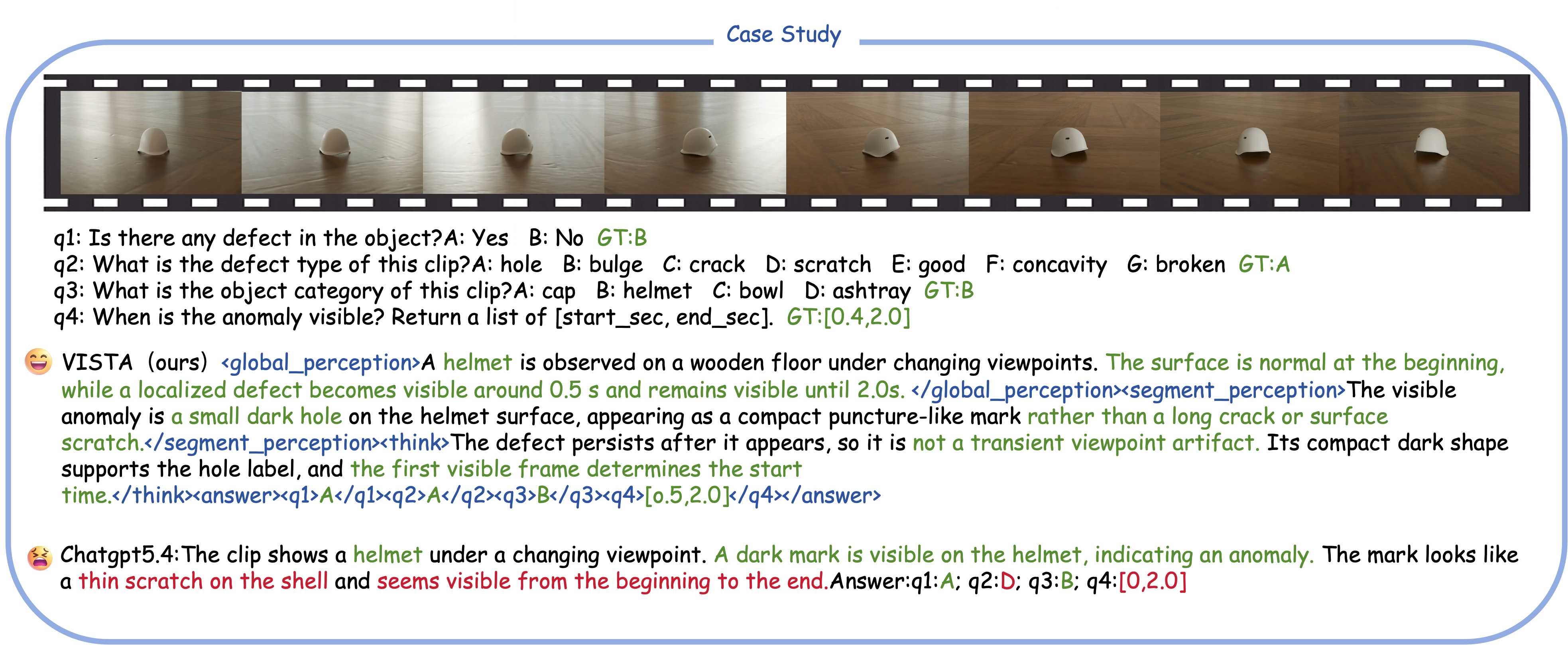}
    \caption{\textbf{Qualitative case study.} VISTA produces perception-structured reasoning that links object-level perception, localized defect evidence, and anomaly visible-time localization.}
    \label{fig:case_study}
\end{figure}

Figure~\ref{fig:case_study} shows a qualitative example of VISTA. The model separates global object perception from localized defect evidence and grounds the defect prediction to the visible-time interval. This illustrates that MMVIAD evaluates not only final answer correctness, but also evidence-grounded temporal understanding.

Table~\ref{tab:training_ablation} compares training paradigms under the same base model on MMVIAD-Unseen. Direct RL without supervised initialization brings limited gains, indicating unstable reward exploration before the model learns structured outputs. PS-SFT substantially improves the base model, and increasing SFT data from 6K to 12K QA pairs further benefits defect classification. The best Avg. is achieved by combining PS-SFT initialization with VISTA-GRPO refinement, which improves the base model from 45.0 to 57.5 and yields a better balance across the four tasks.
Appendix~\ref{app:reward_ablation} further reports additional reward ablations.
\begin{table}[H]
\centering
\caption{\textbf{Training paradigm ablation on MMVIAD.}
We compare different combinations of PS-SFT initialization and VISTA-GRPO refinement. $\mathcal{D}_{\mathrm{PS\text{-}SFT}}$ and $\mathcal{D}_{\mathrm{VISTA\text{-}GRPO}}$ denote the corresponding training data.}
\label{tab:training_ablation}
{\fontsize{7.3}{9.0}\selectfont
\renewcommand{\arraystretch}{0.83}
\resizebox{\textwidth}{!}{%
\begin{tabular}{C{1.05cm} C{1.05cm} C{1.75cm} C{2.15cm} ccc c c}
\toprule
\multirow{2}{*}{PS-SFT} 
& \multirow{2}{*}{VISTA-GRPO} 
& \multirow{2}{*}{$\mathcal{D}_{\mathrm{PS\text{-}SFT}}$} 
& \multirow{2}{*}{$\mathcal{D}_{\mathrm{VISTA\text{-}GRPO}}$}
& \multicolumn{3}{c}{Defect} 
& \multicolumn{1}{c}{Object} 
& \multirow{2}{*}{Avg.} \\
\cmidrule(lr){5-7} \cmidrule(lr){8-8}
& & & & Detect. & Class. & Loc. & Class. & \\
\midrule
\xmark & \xmark & -    & -    & 39.5 & 36.2 & 37.0 & 67.4 & 45.0 \\
\xmark & \cmark & -    & 12k  & 39.6 & 35.5 & 38.3 & 70.0 & 46.0 \\
\cmark & \xmark & 6k   & -    & 53.2 & 37.3 & 46.3 & 81.8 & 54.7 \\
\cmark & \xmark & 12k  & -    & 55.8 & \textbf{42.9} & 48.4 & 75.0 & 55.5 \\
\rowcolor{oursbg}
\cmark & \cmark & 6k   & 6k   & \textbf{60.7} & 37.6 & \textbf{49.6} & \textbf{81.9} & \textbf{57.5} \\
\bottomrule
\end{tabular}
}
}
\end{table}
\section{Conclusion}
We presented MMVIAD, a continuous multi-view video dataset and benchmark for industrial anomaly understanding, with structured labels for defect semantics and anomaly visible-time localization. Built with controllable rendering, MMVIAD enables precise and verifiable temporal annotations that are difficult to obtain from real inspection videos. Our evaluations show that current video MLLMs still struggle with fine-grained defect recognition and temporal grounding, while VISTA improves generalization on MMVIAD-Unseen through PS-SFT initialization and VISTA-GRPO refinement. We discuss limitations and future directions in Appendix~\ref{app:limitations}.

\bibliographystyle{plain}
\bibliography{ref}

\newpage
\appendix
\section{Broader Impact}
\label{app:broader_impact}

\subsection{Potential Positive Impacts}
\label{app:positive_impacts}

Industrial anomaly detection is important for manufacturing quality control, product reliability, and operational safety. By introducing MMVIAD, a continuous multi-view video dataset and benchmark with visible-time annotations, our work aims to support industrial inspection research beyond static-image recognition and toward more realistic video-based anomaly understanding. Compared with single-frame or sparsely sampled settings, MMVIAD evaluates whether models can recognize structural defects under continuous viewpoint changes, understand object semantics, and identify the time interval in which defect evidence is visible. This provides a more diagnostic way to assess whether a model bases its decision on observable visual evidence rather than only producing a final label.

MMVIAD may improve the reliability and interpretability of inspection systems. By coupling anomaly detection, defect classification, object classification, and visible-time localization, the benchmark encourages models to provide both predictions and temporally grounded evidence. Such evidence-grounded outputs can help human inspectors verify model decisions, locate relevant video segments more efficiently, and improve transparency in human-AI collaborative inspection workflows. In practical quality control, this may help reduce missed structural defects, support faster review of suspicious samples, and make model outputs easier to audit.

The benchmark may also benefit multimodal video understanding research. Industrial inspection requires models to attend to fine-grained, local, and structural visual evidence, which is different from many general video understanding tasks. MMVIAD provides a controlled setting for studying fine-grained visual reasoning, temporal grounding, structured outputs, and multi-task consistency. In particular, anomaly visible-time localization asks models to identify when the evidence for a defect is visible, encouraging future models to move from coarse semantic recognition toward more precise evidence localization.

More broadly, MMVIAD may inspire better evaluation protocols for industrial and embodied vision systems. The idea of evaluating whether predictions are temporally aligned with observable evidence may be useful for assembly verification, robotic inspection, equipment monitoring, and dynamic quality control. The structured QA format also provides a unified interface for evaluating detection, recognition, object understanding, and temporal localization, which may help researchers compare model capabilities more systematically across inspection tasks.

MMVIAD may also support reproducibility and fair comparison in industrial anomaly research. Since all tasks share the same input format, structured response protocol, and answer parsing procedure, different models can be evaluated under a unified setting. This can reduce ambiguity in benchmark usage and make it easier to analyze which capabilities are missing, such as fine-grained defect recognition, object-aware reasoning, or temporal evidence grounding. By releasing the benchmark protocol and evaluation format, MMVIAD can provide a common testbed for future studies on industrial video anomaly understanding.

\subsection{Potential Negative Impacts and Risk Mitigation}
\label{app:negative_impacts}

The main risk is over-reliance on automated inspection systems. Models trained or evaluated on MMVIAD should not be treated as deployment-ready safety systems without additional validation in the target industrial environment. In real quality-control workflows, automated predictions should be used to support rather than replace human experts, especially in high-stakes scenarios. Practical deployment should include human oversight, uncertainty monitoring, failure-case analysis, and interpretable evidence such as defect descriptions, key frames, visible-time intervals, or localized visual cues. In addition, inspection videos and model outputs may contain proprietary information about products or manufacturing processes. Deployment should therefore follow appropriate data governance and access-control procedures. With these safeguards, benchmarks such as MMVIAD can support safer, more transparent, and more reliable human-AI collaborative industrial inspection systems.

\section{Reward Design Ablation}
\label{app:reward_ablation}

Table~\ref{tab:reward_ablation} verifies the task-specific effects of our reward design. Removing $R_{\mathrm{sg}}$ mainly reduces defect classification, while replacing $R_{\mathrm{vis}}$ with flat IoU lowers visible-time localization. The full reward design achieves the best Avg., suggesting that semantic gating and visibility-aware temporal modeling provide complementary training signals.

\begin{table}[H]
\centering
\caption{\textbf{Reward design ablation on MMVIAD-Unseen.}
We keep the PS-SFT initialization, VISTA-GRPO training data, and evaluation protocol fixed, and ablate only the reward design during VISTA-GRPO refinement. ``w/o $R_{\mathrm{sg}}$'' removes the semantic gate for defect classification, and ``w/o $R_{\mathrm{vis}}$'' replaces the visibility-aware temporal reward with a flat IoU reward.}
\label{tab:reward_ablation}
{\fontsize{8}{10}\selectfont
\setlength{\tabcolsep}{6pt}
\renewcommand{\arraystretch}{1.05}
\begin{tabular}{c ccc c c}
\toprule
\multirow{2}{*}{Model}
& \multicolumn{3}{c}{Defect}
& Object
& \multirow{2}{*}{Avg.} \\
\cmidrule(lr){2-4} \cmidrule(lr){5-5}
& Detect. & Class. & Loc. & Class. & \\
\midrule
\rowcolor{oursbg}
VISTA-8B
& 60.7 & \textbf{37.6} & \textbf{49.6} & 81.9 & \textbf{57.5} \\
w/o $R_{\mathrm{sg}}$
& \textbf{61.0} & 35.8 & 49.4 & 81.2 & 56.9 \\
w/o $R_{\mathrm{vis}}$
& 59.7 & 37.2 & 48.9 & \textbf{82.1} & 57.0 \\
\bottomrule
\end{tabular}
}
\end{table}

\section{Detailed Information of the MMVIAD}
\label{app:dataset_details}
\subsection{Object Category Taxonomy}
\label{app:object_taxonomy}
This section provides additional details about the object and material composition of MMVIAD. The dataset is constructed from diverse object geometries and rendering factors to support multi-view industrial anomaly understanding under controlled but visually varied inspection conditions.

\begin{table}[H]
\centering
\caption{\textbf{Semantic grouping of object categories in MMVIAD.}
The 48 fine-grained object categories are organized into 17 semantic groups. Train, Test, and Total denote the number of clips in each split and in total, respectively.}
\label{tab:object_category_stats}
{\fontsize{7.5}{10.5}\selectfont
\setlength{\tabcolsep}{2.2pt}
\begin{tabular*}{\textwidth}{@{\extracolsep{\fill}} l c p{7.2cm} c c c}
\toprule
\textbf{Semantic Group} & \textbf{\# Cat.} & \textbf{Fine-grained Categories} & \textbf{Train} & \textbf{Test} & \textbf{Total} \\
\midrule
ashtray & 1 & ashtray0 & 54 & 18 & 72 \\
bottle & 3 & bottle0, bottle1, bottle3 & 183 & 69 & 252 \\
bowl & 6 & bowl0, bowl1, bowl2, bowl3, bowl4, bowl5 & 378 & 129 & 507 \\
bucket & 2 & bucket0, bucket1 & 132 & 54 & 186 \\
cabinet & 1 & cabinet0 & 75 & 15 & 90 \\
cap & 6 & cap0, cap1, cap2, cap3, cap4, cap5 & 360 & 144 & 504 \\
chair & 1 & chair0 & 66 & 24 & 90 \\
cup & 3 & cup0, cup1, cup2 & 162 & 54 & 216 \\
desk & 1 & desk0 & 63 & 21 & 84 \\
eraser & 1 & eraser0 & 54 & 18 & 72 \\
headset & 2 & headset0, headset1 & 108 & 36 & 144 \\
helmet & 4 & helmet0, helmet1, helmet2, helmet3 & 252 & 108 & 360 \\
jar & 1 & jar0 & 54 & 18 & 72 \\
microphone & 2 & microphone0, microphone1 & 111 & 39 & 150 \\
shelf & 1 & shelf0 & 66 & 30 & 96 \\
tap & 2 & tap0, tap1 & 126 & 54 & 180 \\
vase & 11 & vase0, vase1, vase2, vase3, vase4, vase5, vase6, vase7, vase8, vase9, vase10 & 669 & 270 & 939 \\
\midrule
\textbf{Total} & \textbf{48} & - & \textbf{2913} & \textbf{1101} & \textbf{4014} \\
\bottomrule
\end{tabular*}
}
\end{table}
Table~\ref{tab:object_category_stats} summarizes the object taxonomy and split statistics of MMVIAD. The 48 fine-grained object categories are organized into 17 semantic groups, covering diverse object geometries such as containers, furniture, tools, wearable objects, and household items. The train and test splits preserve all semantic groups and fine-grained categories, yielding 2,913 training clips and 1,101 test clips, with 4,014 clips in total. This category coverage provides diverse geometric contexts for evaluating multi-view industrial anomaly understanding.
\begin{figure}[H]
    \centering
    \includegraphics[width=\linewidth]{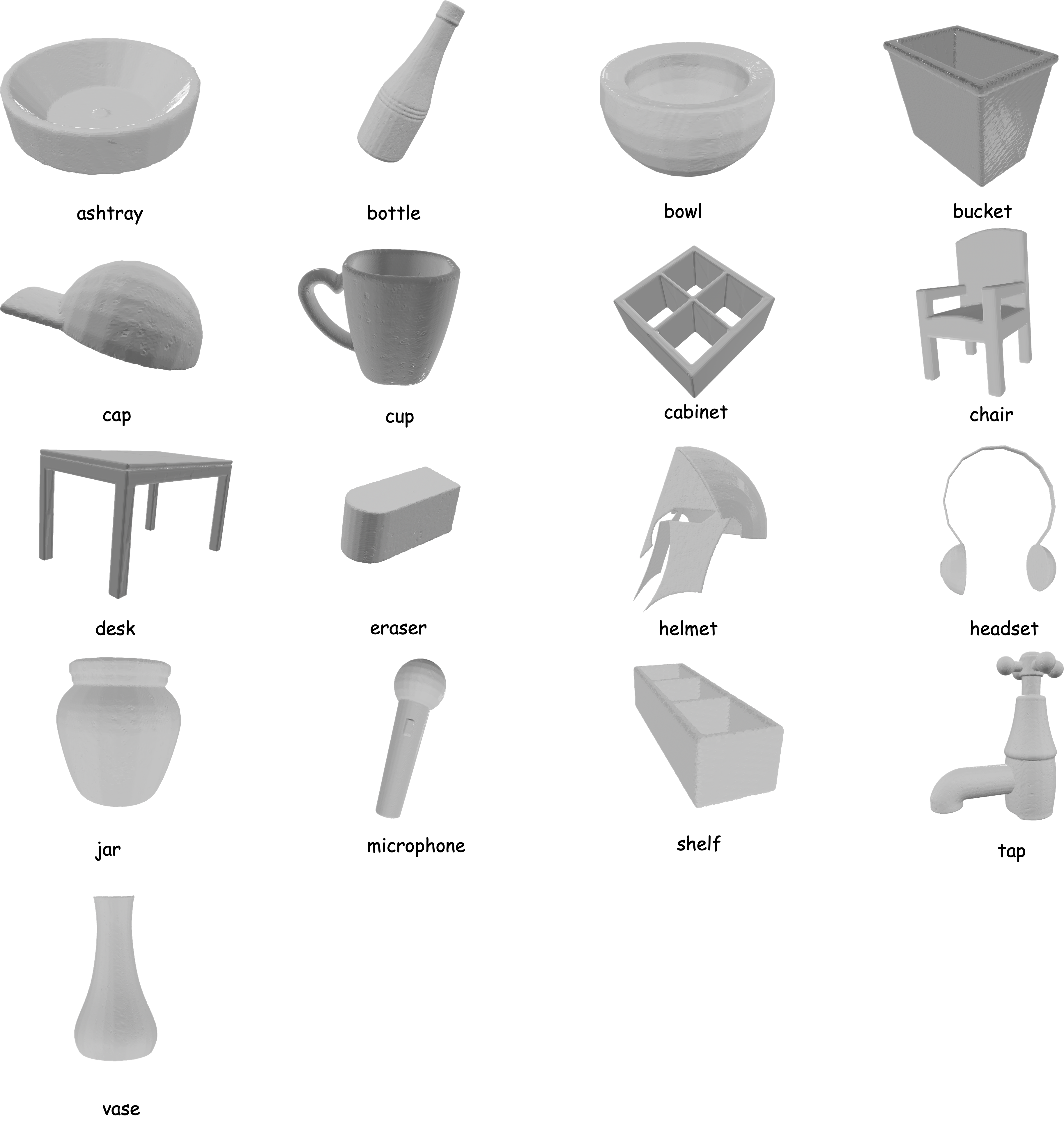}
    \caption{\textbf{Representative semantic object groups in MMVIAD.}
    We visualize representative objects from the 17 semantic groups used in MMVIAD. Objects are rendered with a neutral material to emphasize geometric and semantic diversity.}
    \label{fig:appendix_object_groups}
\end{figure}
MMVIAD contains 48 fine-grained object categories, which are organized into 17 semantic groups. Figure~\ref{fig:appendix_object_groups} visualizes one representative object from each semantic group. These groups cover diverse object shapes and geometric structures for anomaly rendering and inspection.

\subsection{Material Diversity}
\begin{figure}[H]
    \centering
    \includegraphics[width=\linewidth]{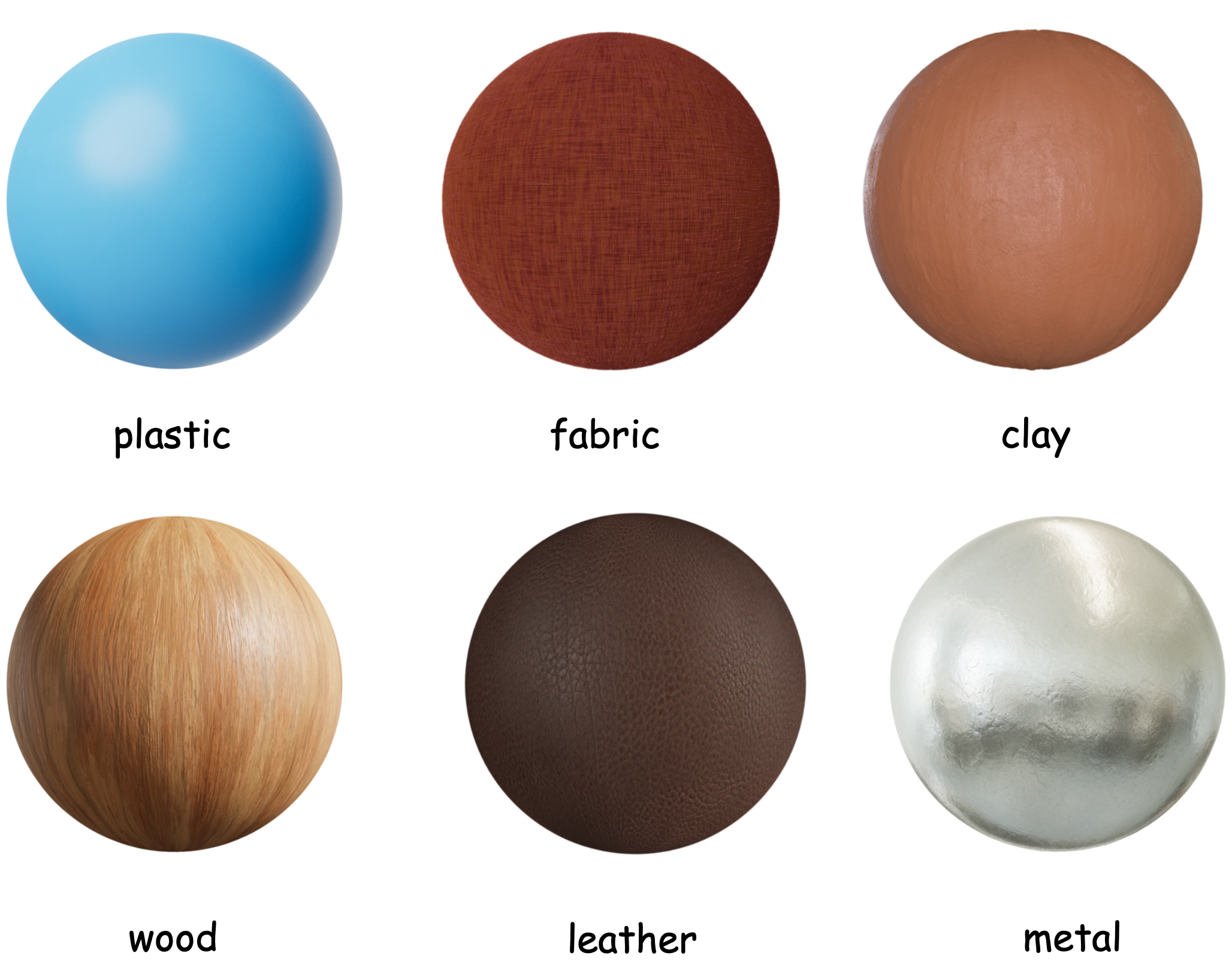}
    \caption{\textbf{Representative material groups in MMVIAD.}
    MMVIAD uses six major material groups, including plastic, fabric, clay, wood, leather, and metal, with 28 fine-grained material subclasses in total. These materials introduce diverse surface appearance and reflectance while preserving the object-centric inspection setting.}
    \label{fig:appendix_material_groups}
\end{figure}

In addition to object-level diversity, MMVIAD introduces material variation during rendering. We organize object materials into six major groups: plastic, fabric, clay, wood, leather, and metal. These groups contain 28 fine-grained material subclasses in total. Figure~\ref{fig:appendix_material_groups} shows representative examples of the six material groups. The material variation changes surface appearance, reflectance, and texture while keeping the underlying object geometry and anomaly structure controlled, allowing MMVIAD to evaluate whether models can recognize structural defects under diverse visual appearances.

Together, the semantic object taxonomy and material variation increase the diversity of MMVIAD beyond simple object identity. The 17 semantic groups provide broad geometric and functional coverage, while the six material groups introduce controlled appearance variation. This design supports evaluation of industrial video anomaly understanding under changes in both object structure and visual surface properties.

\subsection{Structural Anomaly Types}

\begin{figure}[H]
    \centering
    \includegraphics[width=\linewidth]{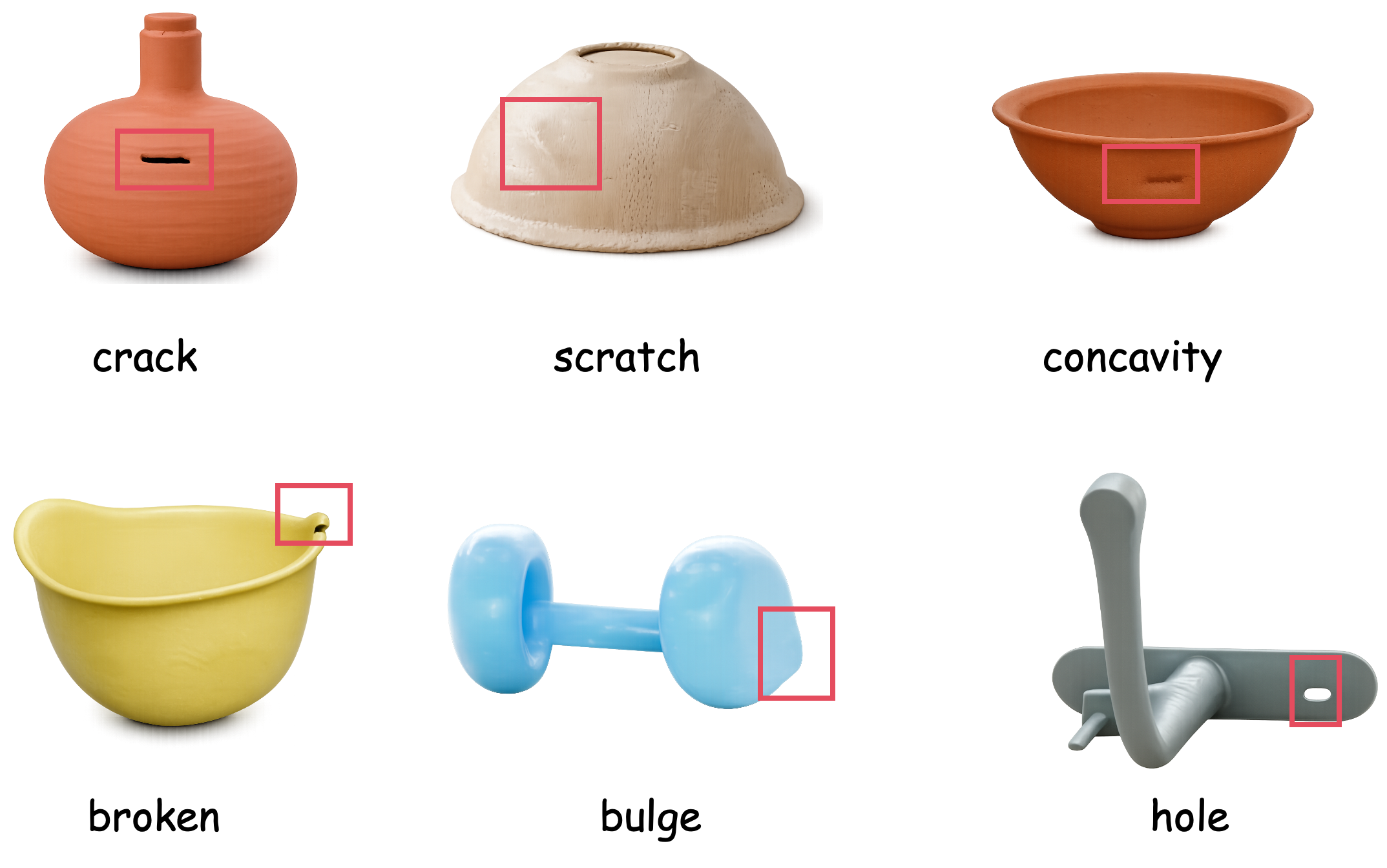}
    \caption{\textbf{Representative structural anomaly types in MMVIAD.}
    We visualize six structural anomaly types, including crack, scratch, concavity, bulge, broken, and hole. The highlighted boxes indicate defect regions on representative objects.}
    \label{fig:appendix_anomaly_types}
\end{figure}

MMVIAD contains six structural anomaly types: crack, scratch, concavity, bulge, broken, and hole. These anomaly types are designed to cover common geometric defects that affect object structure rather than only surface appearance. Figure~\ref{fig:appendix_anomaly_types} shows representative examples of the six anomaly types used in MMVIAD. The highlighted regions indicate the defect locations, illustrating the diversity of defect morphology and visibility across different object shapes.

\section{Reasoning Examples}
\label{app:reasoning_examples}

\begin{figure}[H]
    \centering
    \includegraphics[width=\linewidth]{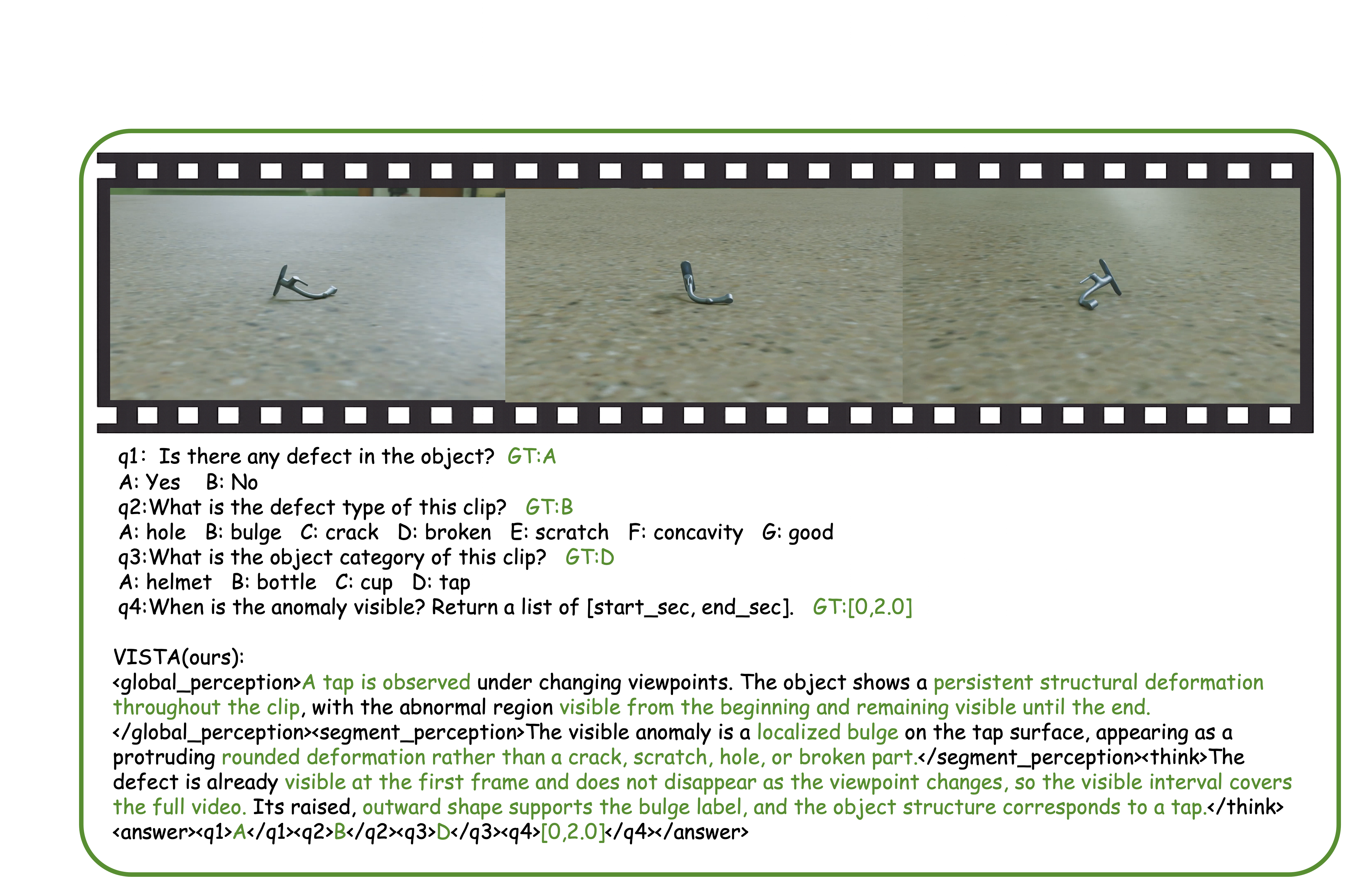}
    \caption{\textbf{Reasoning example: tap with bulge.}}
    \label{fig:appendix_case_tap_bulge}
\end{figure}

\begin{figure}[H]
    \centering
    \includegraphics[width=\linewidth]{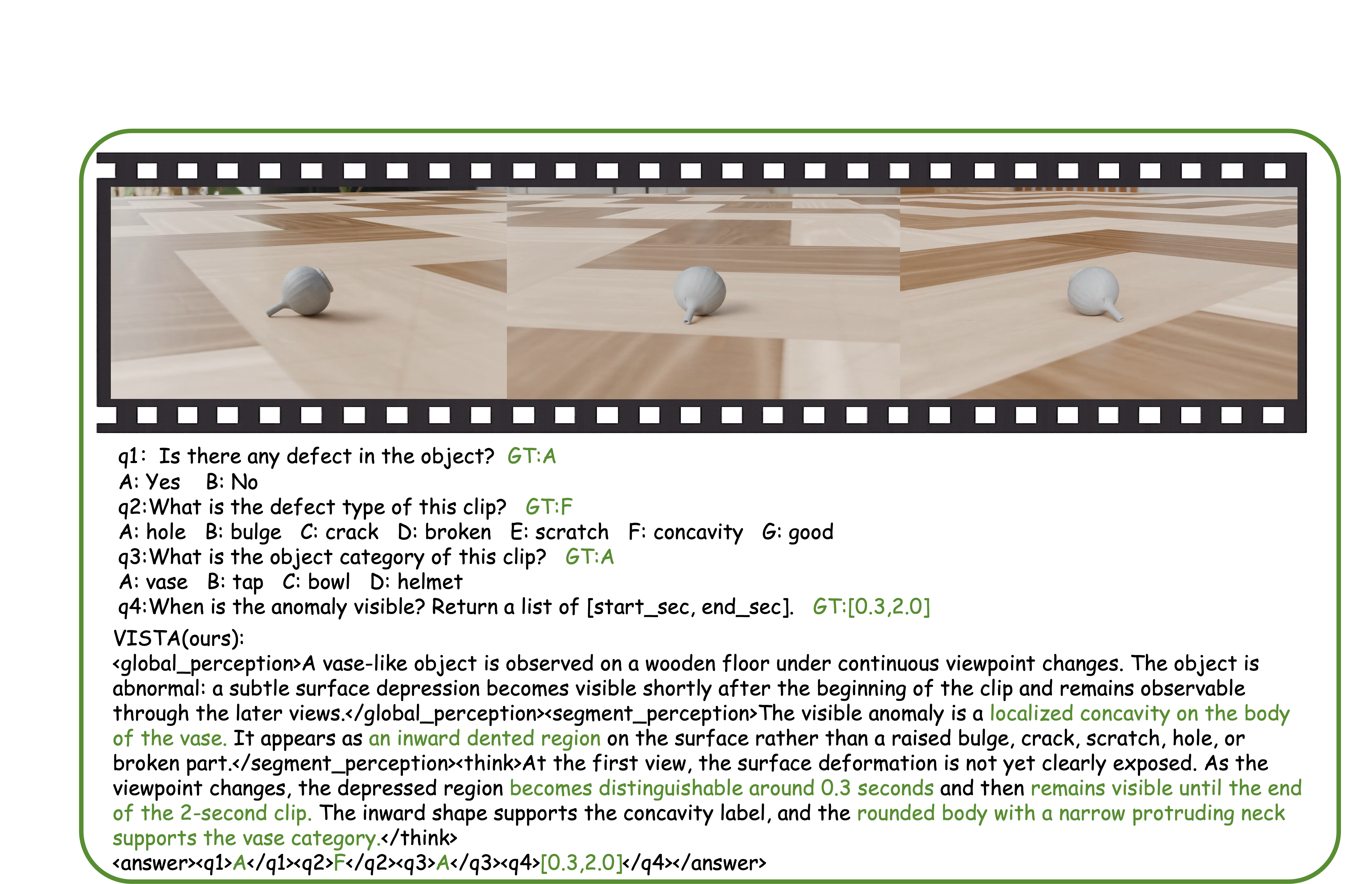}
    \caption{\textbf{Reasoning example: vase with concavity.}}
    \label{fig:appendix_case_vase_concavity}
\end{figure}

\begin{figure}[H]
    \centering
    \includegraphics[width=\linewidth]{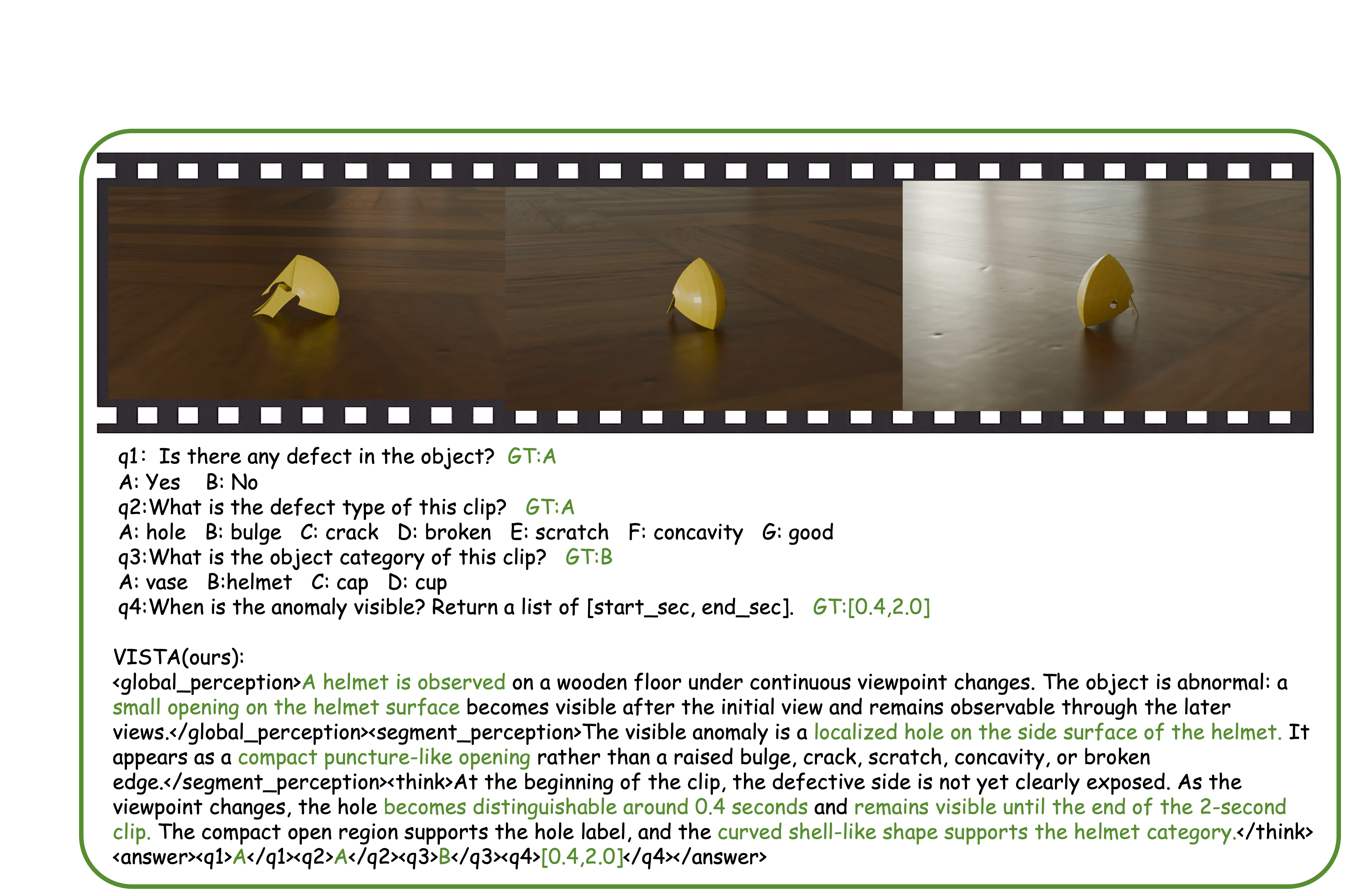}
    \caption{\textbf{Reasoning example: helmet with hole.}}
    \label{fig:appendix_case_helmet_hole}
\end{figure}

\begin{figure}[H]
    \centering
    \includegraphics[width=\linewidth]{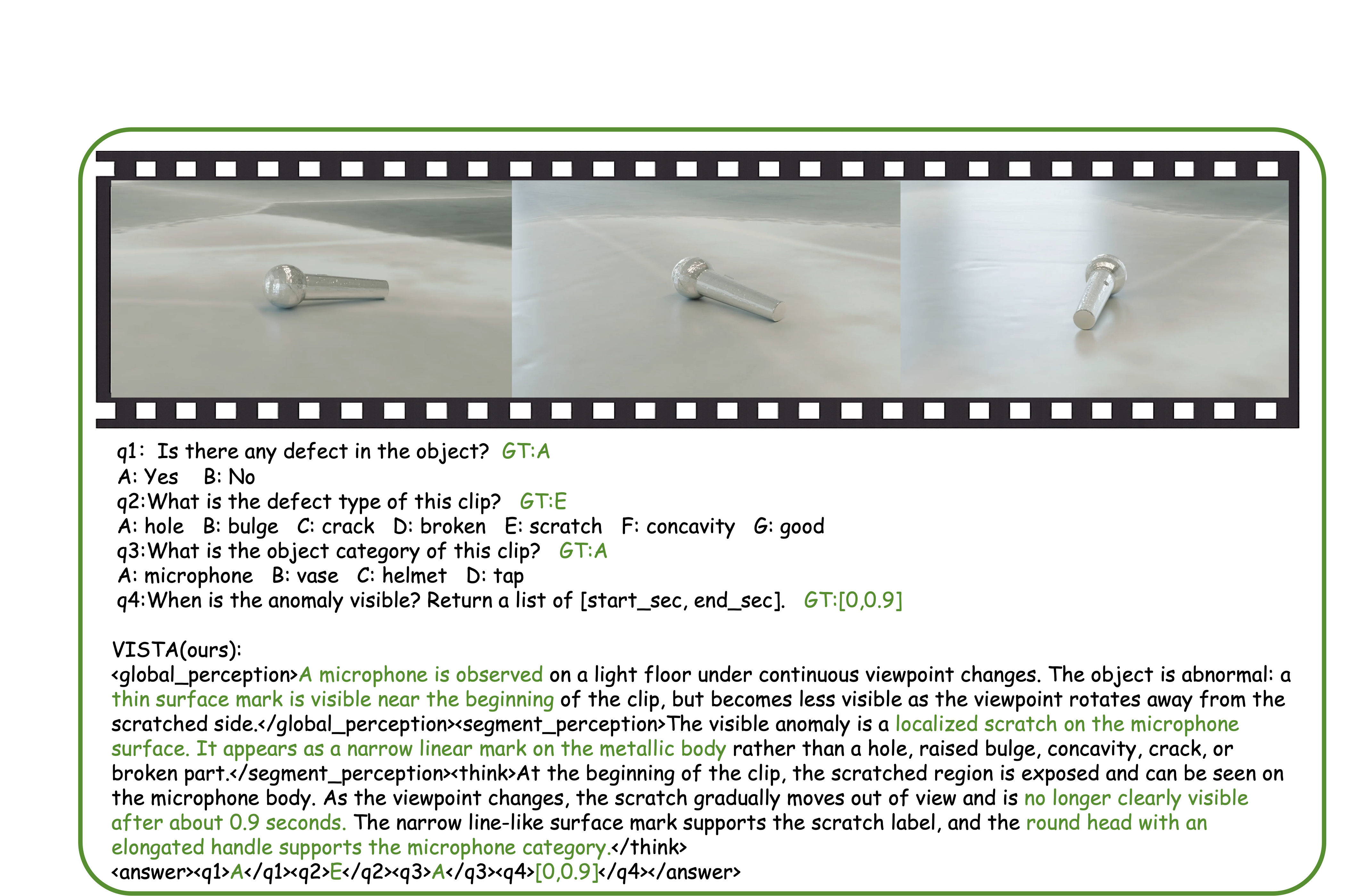}
    \caption{\textbf{Reasoning example: microphone with scratch.}}
    \label{fig:appendix_case_microphone_scratch}
\end{figure}

These examples show that VISTA does not only output final answers, but also grounds
them in explicit visual evidence. The model links object-level cues to category prediction, defect
morphology to anomaly type, and viewpoint-dependent visibility to temporal localization. This
structured reasoning format makes the four task predictions interpretable and exposes whether errors
arise from object recognition, defect semantics, or temporal grounding.
\section{Prompt Template}
\subsection{Main Benchmark Prompt}
\label{app:benchmark_prompt}

For the standard benchmark setting, we use a unified evaluation prompt across all machine baselines. The prompt asks the model to analyze the input video, reason inside \texttt{<think>} tags, and return final predictions in a structured \texttt{<answer>} block. All reported scores are computed only from the parsed final answers.

\begin{lstlisting}
QUESTION_TEMPLATE = (
    "{Question}\n\n"
    "Please think step by step, analyzing the video carefully for any defects or anomalies. "
    "Engage in an internal dialogue using expressions such as 'let me think', 'wait', 'Hmm', "
    "'oh, I see', 'let me verify', etc. "
    "It's encouraged to include self-reflection or verification in the reasoning process. "
    "Provide your detailed reasoning between the <think> </think> tags, and then give your answers "
    "between the <answer> </answer> tags."
)

TYPE_TEMPLATE = (
    "\n\nProvide your answers in the following structured format inside <answer> </answer> tags:\n"
    "<answer>\n"
    "<q1>LETTER</q1>\n"
    "<q2>LETTER</q2>\n"
    "<q3>LETTER</q3>\n"
    "<q4>[[start, end], ...]</q4>\n"
    "</answer>"
)
\end{lstlisting}

\subsection{Structured Generalization Prompt}
\label{app:structured_prompt}

For MMVIAD-Unseen evaluation, we use a structured prompt aligned with the proposed reasoning
protocol. Different from the main benchmark prompt, this prompt requires the model to explicitly
separate whole-clip perception, localized defect evidence, reasoning, and final task predictions. The
final scores are computed only from the parsed \texttt{<q1>}--\texttt{<q4>} fields inside the
\texttt{<answer>} block.

\begin{lstlisting}
STRUCTURED_PROMPT = [
    "You must answer all tasks in a single response using the exact structure below.",
    "You must output all four sections exactly once and in this exact order:",
    "<global_perception> ... </global_perception>",
    "<segment_perception> ... </segment_perception>",
    "<think> ... </think>",
    "<answer> ... </answer>",
    "Do not replace or omit any closing tag.",
    "The <think> section must end with </think>, never </answer>.",
    "The <answer> section must begin with <answer> and end with </answer>.",
    "The final machine-readable result must appear inside <answer>...</answer> only.",
    "Do not put any natural language after <answer> begins; only q1--q4 tags are allowed inside <answer>.",
    "Do not repeat <answer> or any closing tag.",
    "No extra characters are allowed between </think> and <answer>.",
    "Do not output any stray unicode character or separator.",
    "Output exactly this skeleton and fill in content only:",
    "<global_perception>",
    "[content]",
    "</global_perception>",
    "<segment_perception>",
    "[content]",
    "</segment_perception>",
    "<think>",
    "[content]",
    "</think>",
    "<answer>",
    "<q1>LETTER</q1>",
    "<q2>LETTER</q2>",
    "<q3>LETTER</q3>",
    "<q4>[start_sec,end_sec]</q4>",
    "</answer>",
    "",
    "<global_perception>",
    "Describe stable global visual facts: object category clues, material/texture/color/shape, whether the object looks normal or defective overall, and any coarse anomaly morphology. Keep this grounded in the frames only.",
    "</global_perception>",
    "",
    "<segment_perception>",
    "Describe localized defect evidence and temporal visibility cues: anomaly details, which region/part looks abnormal, when it becomes visible in the 2-second clip, and whether it persists or only appears in part of the clip. If no anomaly, state no anomaly-specific visible segment and no localized defect evidence.",
    "</segment_perception>",
    "",
    "<think>",
    "Reason briefly from the visual perception to the final answers for all tasks. Do not introduce facts not supported by the visual perception.",
    "</think>",
    "",
    "<answer>",
    "<q1>A or B</q1>",
    "<q2>LETTER</q2>",
    "<q3>LETTER</q3>",
    "<q4>[start_sec,end_sec]</q4>",
    "</answer>",
    "",
    "Additional constraints:",
    "- <global_perception> must be 80 words or fewer.",
    "- <segment_perception> must be 80 words or fewer.",
    "- <think> must be 60 words or fewer.",
    "- The final machine-readable result must appear only inside <answer>...</answer>.",
    "- Do not wrap the answer in markdown fences.",
    "- anomaly_detection must be encoded in <q1> as A or B.",
    "- defect_classification must be encoded in <q2> as the option letter.",
    "- object_classification must be encoded in <q3> as the option letter.",
    "- visible_time_localization must be encoded in <q4> as [] or [start_sec,end_sec].",
    "- For visible_time_localization, use the earliest and latest times when the anomaly is visible at all, even if subtle.",
    "- Do not delay the start time to only when the anomaly becomes most obvious.",
    "- Keep <global_perception> and <segment_perception> concise and clean.",
    "- The parser will only use the q1--q4 tags inside <answer>...</answer> as final prediction.",
]
\end{lstlisting}

\subsection{Single-View Image Generalization Prompt}
\label{app:single_view_prompt}

For the single-view image generalization evaluation on MVTec AD and VisA, we adapt the structured prompting protocol used in MMVIAD-Unseen to the image setting. Although these datasets contain static images rather than videos, we keep the same perception-structured response format to reduce prompt-format differences. The prompt asks the model to determine whether a defect is present in the image, and all reported scores are computed only from the parsed final answer.

\begin{lstlisting}
SINGLE_VIEW_PROMPT = """
Look at the image carefully and answer the following question about object anomaly detection.

The object in this image is: {category}

Q1: Is there a defect or anomaly present in this image?
Options: (A) Yes, a defect is present. (B) No, the object looks normal.

Think step by step using the structured format below.

<global_perception>
[Describe the object's overall appearance: material, texture, color, shape, and whether it looks normal or defective. <=80 words.]
</global_perception>

<segment_perception>
[Describe any localized defect evidence: which region looks abnormal. If no anomaly, state: no anomaly visible. <=80 words.]
</segment_perception>

<reasoning>
[Reason briefly to your final answer. <=60 words.]
</reasoning>

<answer>
{"anomaly_detection": "A or B"}
</answer>
"""
\end{lstlisting}

\section{Additional Training Curves}
\label{app:training_curves}

\begin{figure}[H]
    \centering
    \includegraphics[width=\textwidth]{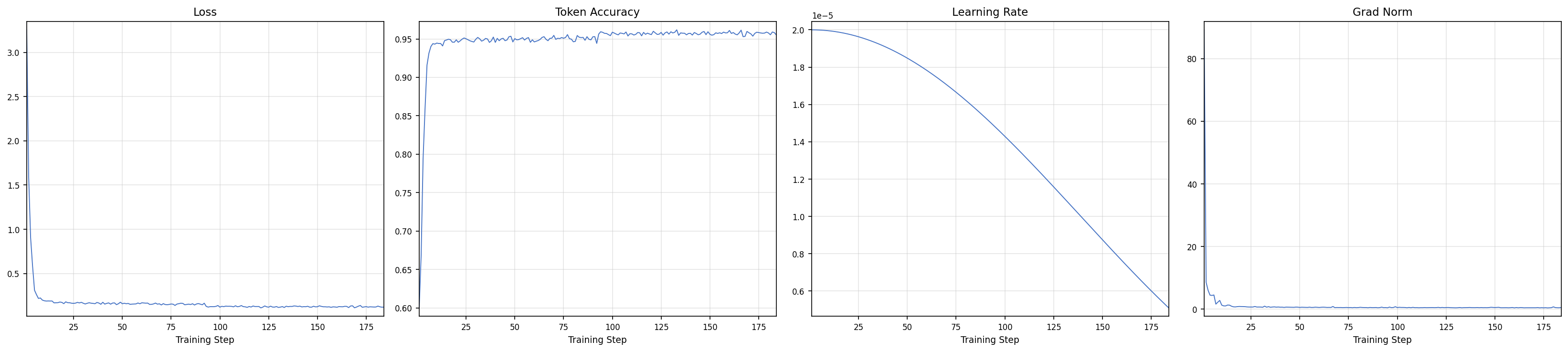}
    \caption{\textbf{PS-SFT training curves.} We report loss, token accuracy, learning rate, and gradient norm during perception-structured supervised fine-tuning.}
    \label{fig:sft_training_curves}
\end{figure}

\begin{figure}[H]
    \centering
    \includegraphics[width=\textwidth]{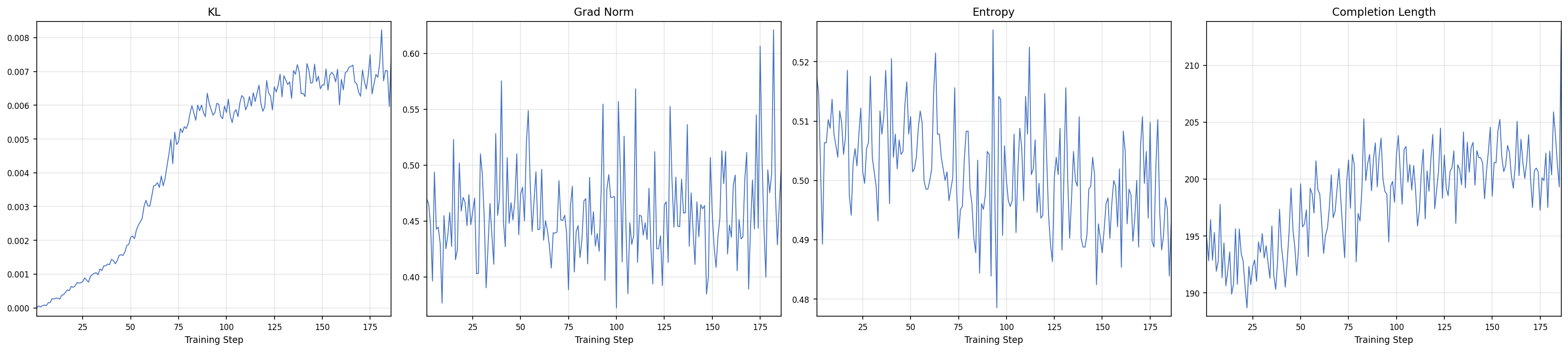}
    \caption{\textbf{VISTA-GRPO training curves.} We report KL divergence, gradient norm, entropy, and completion length during reinforcement learning refinement.}
    \label{fig:rl_training_curves}
\end{figure}

This section provides additional training dynamics for the two-stage post-training pipeline. Figure~\ref{fig:sft_training_curves} shows the PS-SFT training curves, including loss, token accuracy, learning rate, and gradient norm. Figure~\ref{fig:rl_training_curves} shows the VISTA-GRPO training curves, including KL divergence, gradient norm, entropy, and completion length.

\section{Limitations and Future Work}
\label{app:limitations}

MMVIAD takes a first step toward continuous multi-view video understanding for industrial anomaly detection, but it has several limitations. First, the dataset scale is still moderate. MMVIAD is built from high-quality object-centric 3D assets that support 360$^\circ$ rendering and controllable structural anomaly synthesis, but such assets are currently limited in availability. As a result, expanding the dataset to more object categories, defect types, and industrial scenarios remains an important direction. Second, MMVIAD is synthetic. Although the rendering process introduces diverse viewpoints, materials, floors, and HDRI environments, there is still a domain gap between rendered inspection videos and real industrial acquisition, where sensor noise, motion blur, imperfect lighting, occlusion, and production-line constraints may appear. Future work should combine synthetic generation with real captured multi-view inspection videos to improve realism and deployment relevance.

Another limitation is the computational cost of data generation. MMVIAD uses high-resolution rendering and dense frame rates to preserve fine structural details and temporally localized anomaly evidence. This improves annotation quality and supports visible-time localization, but also makes rendering time-consuming and expensive. More efficient rendering pipelines, adaptive frame sampling, and targeted generation of difficult viewpoints could make future dataset construction more scalable. In addition, the current benchmark focuses on short 2-second clips with approximately 120$^\circ$ viewpoint changes. This design provides a compact and controlled setting, but longer inspection trajectories, multi-object scenes, and more complex camera-object motion would better reflect real-world inspection workflows.

Finally, VISTA-GRPO improves structured anomaly understanding, but fine-grained defect classification remains challenging, especially under sparse and imbalanced defect distributions. Future work may explore denser semantic rewards, hierarchical defect taxonomies, stronger temporal grounding objectives, and larger multimodal backbones. We also plan to study how MMVIAD can support more realistic inspection protocols, including open-vocabulary defect discovery, uncertainty estimation, and interactive human-AI verification for safety-critical industrial inspection.


\end{document}